\documentclass{article} % For LaTeX2e
\usepackage{iclr2026_conference,times}

\usepackage{amsmath,amsfonts,bm}

\def\Figref#1{Figure~\ref{#1}}

\def\eqref#1{equation~\ref{#1}}
\def\1{\bm{1}}

\DeclareMathAlphabet{\mathsfit}{\encodingdefault}{\sfdefault}{m}{sl}
\SetMathAlphabet{\mathsfit}{bold}{\encodingdefault}{\sfdefault}{bx}{n}

\def\sR{{\mathbb{R}}}

\usepackage{booktabs}
\usepackage{array}
\usepackage{multirow}
\usepackage{makecell}
\usepackage{colortbl}
\usepackage{hyperref}
\usepackage{url}
\usepackage{adjustbox}
\usepackage{graphicx}
\usepackage[T1]{fontenc}
\usepackage{subcaption}
\usepackage{fontawesome}
\usepackage{enumitem}
\usepackage{wrapfig}
\usepackage{xcolor}
\usepackage[most]{tcolorbox}
\definecolor{ggreen}{rgb}{0.0, 0.6, 0.0}
\definecolor{rred}{rgb}{0.75, 0.0, 0.0}
\definecolor{bblue}{rgb}{0.13, 0.67, 0.8}

\definecolor{BoxBackground}{RGB}{240, 240, 240} % 浅灰色背景
\definecolor{BoxFrame}{RGB}{0, 0, 0} % 黑色边框
\definecolor{TitleBackground}{RGB}{0, 0, 0} % 标题背景颜色
\definecolor{TitleText}{RGB}{255, 255, 255} % 标题文字颜色
\definecolor{deepgreen}{RGB}{0,100,0}
\tcbset{
  academicbox/.style={
    boxsep=5pt,
    left=2pt,
    right=2pt,
    bottom=0.5pt,
    boxrule=0.5pt,
    colback=BoxBackground,
    colframe=BoxFrame,
    colbacktitle=TitleBackground,
    coltitle=TitleText,
    enhanced,
    attach boxed title to top left={yshift=-0.1in,xshift=0.1in},
    boxed title style={boxrule=0pt,colframe=white},
    title={#1},
  }
}

\newtcolorbox{AcademicBox}[1][]{academicbox=#1}
\definecolor{SoftBlue}{RGB}{135, 206, 250}  % 浅蓝色
\definecolor{SoftOrange}{RGB}{255, 224, 178} % 浅橙色
\definecolor{SoftGreen}{RGB}{144, 238, 144}  % 浅绿色
\definecolor{CorrectGreen}{RGB}{76, 175, 80} % 淡绿色，适用于表示正确
\definecolor{ErrorRed}{RGB}{211, 47, 47} % 深红色，适用于表示错误

\usepackage{graphicx}

\title{Revisiting Long-context Modeling from Context Denoising Perspective}

\author{Zecheng Tang$^{1,2}$,\quad Baibei Ji$^{1,2}$,\quad Juntao Li$^{1,2}$\thanks{Corresponding Author},\quad Lijun Wu$^{3}$,\quad Haijia Gui$^{1}$,\quad Min Zhang$^{1}$ \\
$^{1}$Soochow University \quad $^{2}$ LCM Laboratory \quad $^{3}$Shanghai Artificial Intelligence Laboratory \\
\texttt{\{zctang, bbji\}@stu.suda.edu.cn} \quad  \texttt{\{ljt, minzhang\}@suda.edu.cn}
}

\iclrfinalcopy % Uncomment for camera-ready version, but NOT for submission.
\begin{document}

\maketitle
\vspace{-1em}
\begin{center}
    \textbf{\texttt{\faGithub~Code: \textcolor{violet}{ \url{https://github.com/LCM-Lab/context-denoising-training}}}}
\end{center}
\vspace{1em}

\begin{abstract}
Long-context models~(LCMs) have demonstrated great potential in processing long sequences, facilitating many real-world applications.
The success of LCMs can be attributed to their ability to locate implicit critical information within the context for further prediction. 
However, recent research reveals that LCMs are often susceptible to contextual noise, i.e., irrelevant tokens, that can mislead model attention.
In this paper, we conduct a fine-grained analysis of the context noise and propose an effective metric, the Integrated Gradient~(IG) score, to detect and quantify the noise information within the context. 
Our findings reveal that even simple mitigation of detected context noise can substantially boost the model's attention on critical tokens and benefit subsequent predictions.
Building on this insight, we propose Context Denoising Training~(CDT), a straightforward yet effective training strategy that improves attention on critical tokens while reinforcing their influence on model predictions.
Extensive experiments across four tasks, under both context window scaling and long-context alignment settings, demonstrate the superiority of CDT.
Notably, when trained with CDT, an open-source 8B model can achieve performance~(50.92) comparable to GPT-4o~(51.00).
% \footnote{Code:~\url{https://anonymous.4open.science/r/context-denoising-training-D7DF}}.
\end{abstract}

\section{Introduction}
\label{sec:introduction}

\begin{wrapfigure}{r}{0.5\linewidth}
    \centering
    \vspace{-17pt}
    \includegraphics[width=\linewidth]{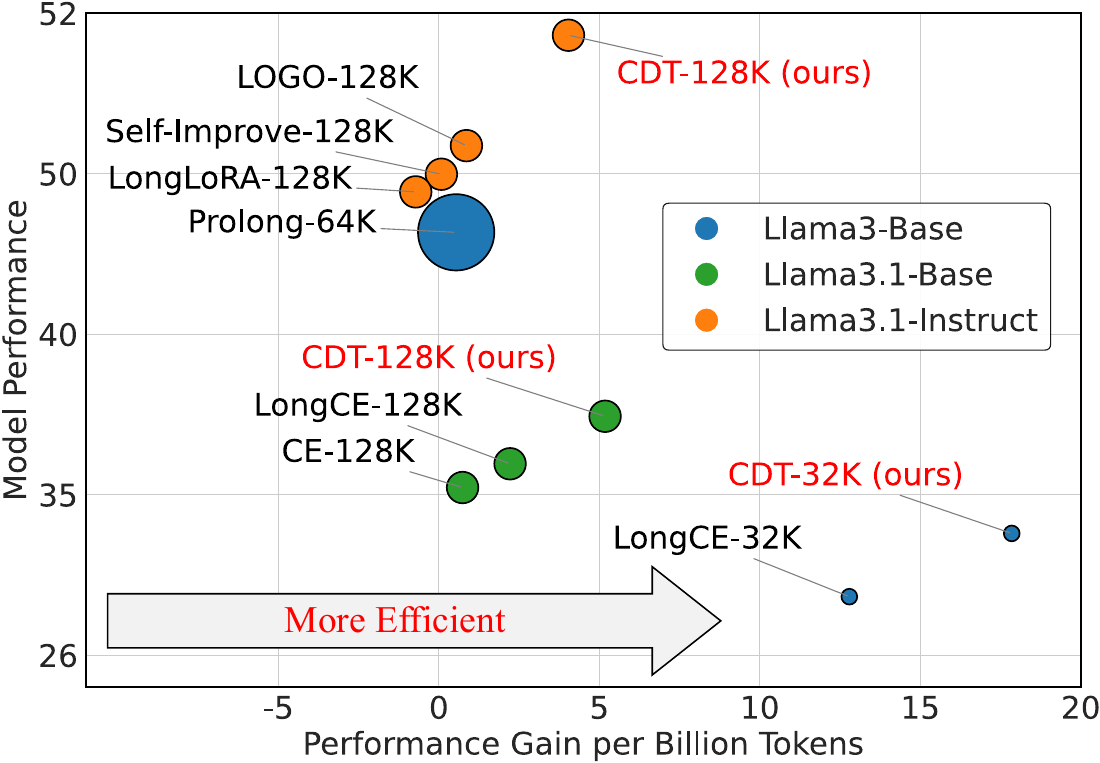}
    \caption{Comparative overview of \textit{model performance} on real-world long-context tasks and \textit{performance gain per billion tokens} among different training methods. The bubble size indicates the relative training data volume.}
    \label{fig:intro}
    \vspace{-10pt}
\end{wrapfigure}

The ability to handle long input sequences has become a fundamental requirement for large language models~(LLMs), with cutting-edge models capable of processing context lengths exceeding millions of tokens~\citep{team2024gemini,minimax2025minimax01scalingfoundationmodels,meta2025llama,qiu2025gated}. 
This advancement eliminates the need for complex toolchains and intricate workflows, e.g., RAG~\citep{yu2024defense}, and significantly enhances real-world applications, such as LLM agent~\citep{luo2025large,xi2025agentgym} and project code analysis~\citep{fang2024large}.

Recent studies indicate that LCMs frequently fail when processing long-context tasks~\citep{hsieh2024ruler,kuratov2024babilong,tang2024citeeval,bai2024longbench2}, and the open-source community mitigates such an issue mainly by using sufficient high-quality synthetic long-context data to post-train the model~\citep{fudata,chen2024essential,gao2024quest}. 
However, these approaches are proven to be either inefficient or ineffective under limited resources~(Appendix~\ref{appdix:train_efficiency}).
For example, as shown in \Figref{fig:intro}, Prolong-64K-Base~\citep{gao2024prolong} achieves significant performance but improves by only 0.3 points per 1B tokens used. 
In contrast, LongCE~\citep{fang2024wrong} exhibits less improvement but achieves nearly 13 points per 1B tokens, demonstrating significantly higher training efficiency.

One of the possible reasons is that existing works overlook the fact that LCMs process long input in an implicit \emph{retrieval-then-generation} manner, i.e., first identifying key information within the context and then further generating with the ``retrieved-context''~\citep{liu2024lost,wu2024retrieval,li2024alr,qiu2025eliciting}.
However, the critical tokens in the ``retrieved-context'' might be overwhelmed by excessive irrelevant tokens~\citep{ye2024differential}.
Thus, the key to achieving better long-context modeling is \emph{effectively detecting the critical tokens, diminishing the effect of irrelevant tokens~(context noise), and strengthening the connection between model prediction and critical tokens.}
Conventional language modeling training strategy, which relies on uniform token-wise supervision through cross-entropy loss, is fundamentally inefficient for long-context modeling because it cannot distinguish critical tokens from irrelevant tokens in lengthy inputs.

In this work, we first investigate the impact of context noise on long-context modeling. 
Specifically, we propose a novel critical token detection metric, the Integrated Gradient (IG) score, based on the concept of information flow~\citep{wang2023label}. 
Our approach achieves a remarkable accuracy improvement in the critical token detection task compared to the traditional attention-based method.
Then, we leverage the IG score to manually reduce the context noise by subtracting the gradient values associated with irrelevant tokens from the token embeddings.
We find that simply suppressing context noise at the model input allows LCMs to focus more effectively on critical tokens.

Built upon the above analysis, we further propose a simple yet effective Context Denoising Training~(CDT) strategy, which performs denoising at the model input, allowing the model to focus more effectively on critical tokens to better establish the connection between critical tokens and generation.
Notably, our CDT approach is analogous to the \textit{Signal Denoising} in the digital signal processing field~\citep{kopsinis2009development}, where noise reduction in the input sequence can enhance the model's attention to essential parts within the context.
Experiments on two essential long-context training scenarios, i.e., context window scaling and long-context alignments, across 4 different types of long-context tasks (real-world tasks, language modeling task, synthetic tasks, and long-form reasoning tasks) exhibit the superiority of our method.
Our CDT can consistently surpass the other methods with an average gain of 2 points on 12 real-world long-context tasks in LongBench-E~\cite{bai2024longbench} and 13 long synthetic tasks in RULER~\citep{hsieh2024ruler}.
Additionally, with CDT, an open-source Llama3.1-8B-Instruct model can achieve comparable results with GPT4o on real-world tasks~(50.92 points v.s. 51.00 points on LongBench-E testing set).

% In summary, our contributions are:
% \begin{itemize}[itemsep=2pt,topsep=0pt,parsep=0pt,leftmargin=1.5em]
% \item We introduce the IG score, which can more effectively identify critical tokens in long contexts than standard attention scores. Additionally, we derive a theoretical approximation of the IG score using token embedding gradients.
% \item We propose a simple yet effective training strategy for context denoising, i.e., CDT, which demonstrates strong performance across two long-context training settings and four task types.
% \item With our method, an open-source 8B model achieves performance on real-world tasks (50.92 points) that is comparable to GPT-4o (51.00 points).
% \end{itemize}
\vspace{-0.5em}

\section{Related Work}  
\label{sec:related_work}

\subsection{Retrieval-then-generation Mechanism of Long-context Models}
Existing research has demonstrated that LCMs handle long-context in a ``retrieval-then-generation'' manner, where \emph{LCMs first retrieve salient information within the context and utilize this information for further prediction}~\citep{wu2024retrieval,tang2024citeeval,zhao2024understanding,qiu2025eliciting}.
However, \citet{liu2024lost} observes the ``lost-in-the-middle'' phenomenon of LCMs, which highlights that LCMs exhibit a positional bias toward locating key information.
Furthermore, \citet{ye2024differential} and \citet{fang2024wrong} discover that excessive irrelevant long-context can overwhelm critical information, thereby impairing the performance of the model.
To mitigate the above issue, some works have explored solutions from various perspectives, including model architecture improvements~\citep{ye2024differential,xiao2024duoattention}, enhancements in information extraction mechanisms~\citep{li2024alr,zhang2024longcite}, and optimization of training objective~\citep{fang2024wrong,bai2024longalign}.
In this paper, we revisit critical information location from the context denoising aspect, helping the model establish better connections between detected salient tokens and predictions.

\subsection{Long-context Post-training}
Generally, the purposes of long-context post-training can be categorized into two types: \emph{context window scaling} and \emph{long-context alignment}.
For context window scaling, prior studies have managed to extend the context length of LLMs with limited computational cost compared to pretraining.
It can be further categorized into two approaches: positional extrapolation\citep{chen2023extending,pengyarn,ding2024longrope,liu20242,zhao2024longskywork,zhang2024extending,fudata2024,lu2024controlled,wang2025layer,ge2025bytescale} and model architecture modification\citep{chevalier2023adapting,chen2023longlora,xiaoefficient,bertsch2024unlimiformer,yuan2025native,lu2025moba}.
Another line of work focuses on improving models that already support long-context windows, aiming to enhance the model’s ability to capture critical information from lengthy contexts~\citep{liu2024lost,an2024make,gao2024train,an2024does} and to address alignment challenges such as hallucination~\citep{zhang2024longreward,tang2024logo,li2024large}.
However, to date, no existing work has simultaneously considered both training efficiency and effectiveness under the two aforementioned settings.
Only a few preliminary studies~\citep{fang2024wrong,helm2025token} have explored token re-weighting or critical token pruning~\citep{guo2025learning} as a trivial method to achieve a trade-off between long-context processing efficiency and effectiveness.

\section{Preliminary Study} % Locating Salient Tokens and Diminishing Noise in Long Context

\begin{figure*}
    \centering
    \includegraphics[width=\linewidth]{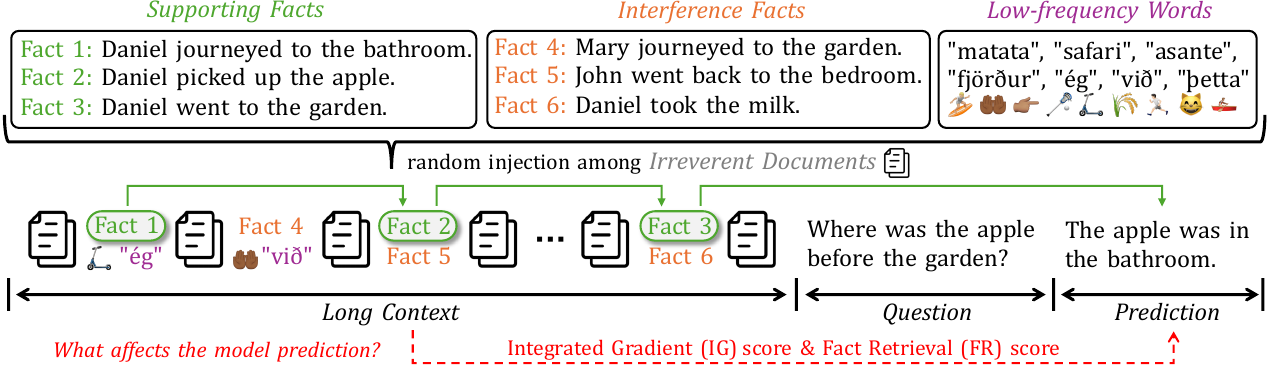}
    \caption{Task format of our preliminary study, which requires models to predict the final answer by reasoning through multi-hop Supporting Facts and distinguishing from the Interference Facts. Simultaneously, the model should also resist the influence of Irreverent Documents and Low-Frequency Words. More details are shown in Appendix~\ref{appdix:pre_task}.}
    \label{fig:pre_task_format}
\end{figure*}

\label{sec:preliminary}
In this section, we analyze the influence of context noise, i.e., irrelevant tokens, on long-context modeling.
More concretely, we first design critical token detection metrics in \S\ref{pre:sub1} and study the impact of context noise restraint on long-context modeling in \S\ref{pre:sub2}.
For evaluation, we construct a synthetic long-form reasoning task as a controlled proxy to enable precise assessment, due to the lack of real-world testing data with explicitly labeled critical token positions.
We conduct experiments with the Llama3.1-8B-Instruct~\citep{meta2024introducing} model, which owns a 128K context window size.

\paragraph{Synthetic Task Format}
As shown in Figure~\ref{fig:pre_task_format}, there are four types of tokens in the context: supporting facts, interference facts, low-frequency words, and irrelevant documents.
The model’s task is to predict the correct answer (e.g., ``bathroom'') by reasoning over \textcolor{green}{supporting facts}.
The \textcolor{orange}{interference facts} are seemingly related to the answers and are randomly inserted into the context, aiming to distract the models from providing the correct response. 
We treat both supporting facts and interference facts as \emph{critical tokens}, as they are both highly correlated with the answer.
The key distinction lies in semantic validity: LCMs must discern which tokens are genuinely supportive — and which are misleading — to predict accurately.
Besides, models should also prevent critical tokens from being overwhelmed by \emph{irrelevant tokens}, including excessive \textcolor{gray}{irrelevant documents} and \textcolor{violet}{low-frequency words}.
The total context length of each sample ranges from 0K to 64K.

\subsection{Critical Tokens Detection}
\label{pre:sub1}
Given the model input $X=\{x_i\}_{i=1}^{n}$ which contains $n$ tokens and the ground truth $Y=\{y_j\}_{j=1}^{m}$ which contains $m$ tokens, we design two metrics to reflect the influence of context noise: Fact Retrieval~(FR) score and Integrated Gradient~(IG) score.

\paragraph{Attention Distribution Metric: FR score}
Existing works primarily identify critical tokens based on the attention distribution~\citep{wu2024retrieval,gema2024decore,xiao2024duoattention}. 
Similarly, we design the Fact Retrieval~(FR) score for our synthetic task based on the attention distribution to quantify the model's attention allocated to different types of tokens.
At each step of model prediction $y_j$, if the attention score of $x_i$ ranks within the top-k across the entire sequence, we define $x_i$ as being attended by an attention head $h$ in the $l$-th model layer.
Let $s_j$ be the set of tokens attended by an attention head $h$ at the generation step $j$, and $\mathcal{T}_{r}$ refers to the context token set of type $r\in \{\mathrm{sup, inter, irr, low}\}$, e.g., $\mathcal{T}_{sup}$ denotes tokens of the supporting facts. 
The FR score $\mathrm{FR}^{(r)}_{h,l}$ of the $h$-th attention head in the $l$-th model layer can be written as:
\begin{align}
    \mathrm{FR}^{(r)}_{h,l} = \frac{\mid s_j \cap \mathcal{T}_{r}\mid}{\mid \mathcal{T}_{r}\mid}.
    \nonumber
\end{align}
We average FR scores from all heads to reflect the attention distribution of tokens in $\mathcal{T}_{r}$.

\begin{figure}[t]
    \centering
    \begin{subfigure}[b]{0.49\linewidth}
        \centering
        \includegraphics[width=\linewidth]{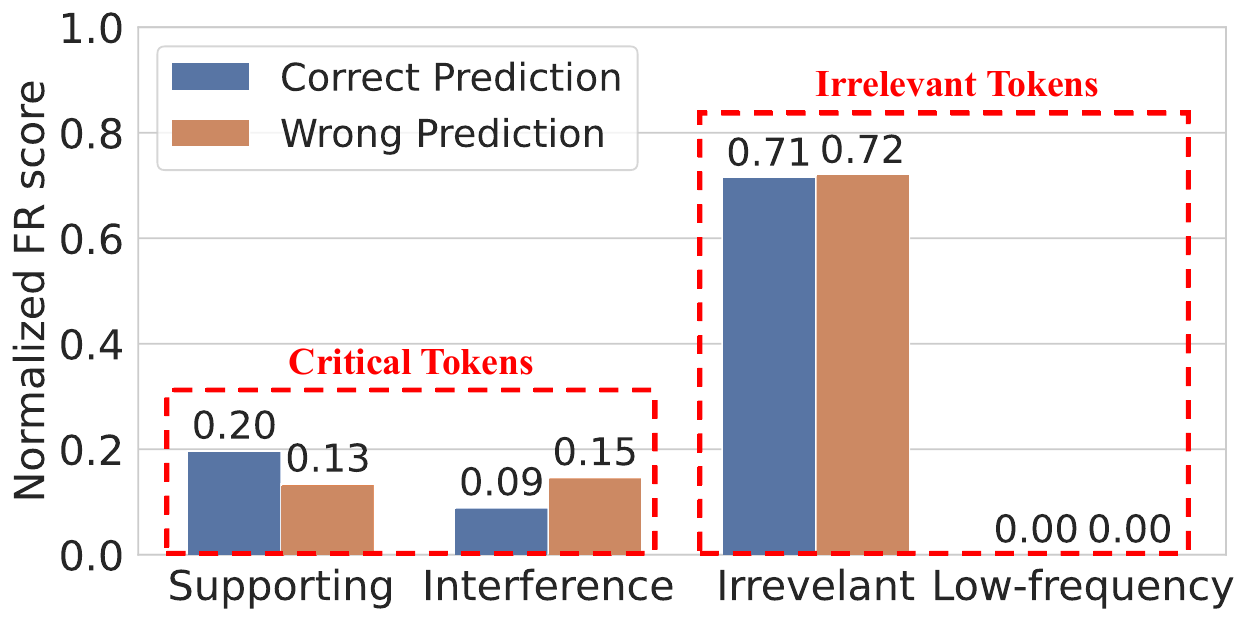}
        \caption{Attention distribution reflected by FR score.}
        \label{fig:pre1_attn}
    \end{subfigure}
    \hfill
    \begin{subfigure}[b]{0.49\linewidth}
        \centering
        \includegraphics[width=\linewidth]{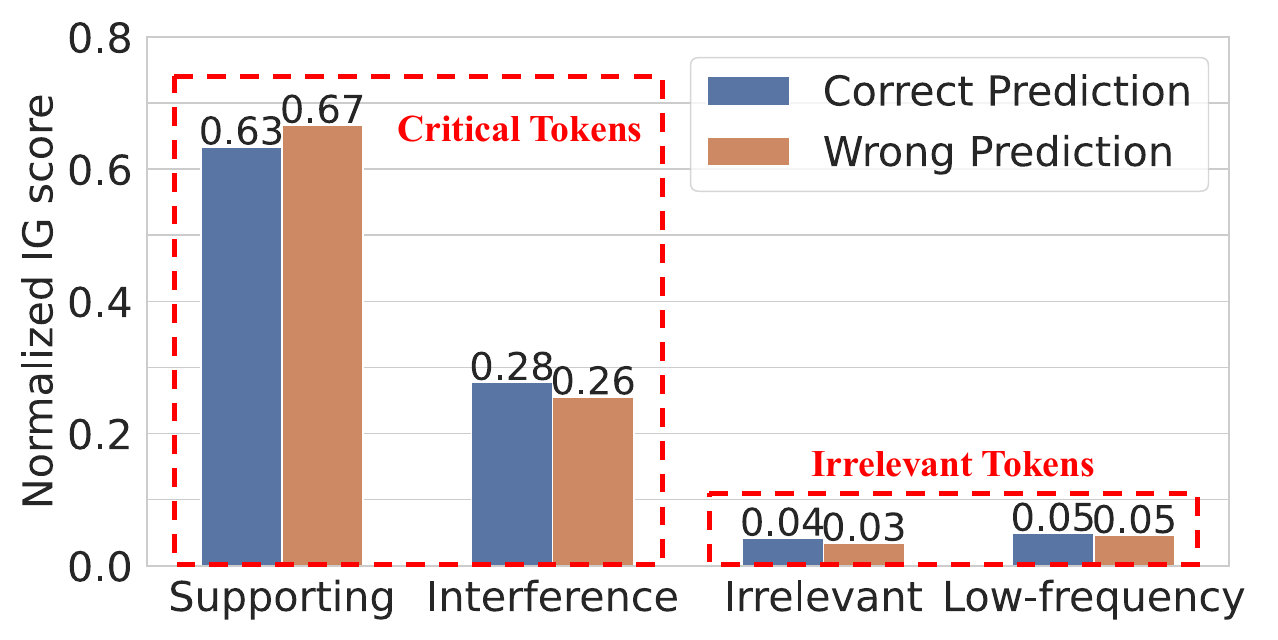}
        \caption{Information flow reflected by average IG score.}
        \label{fig:pre1_infor}
    \end{subfigure}
    \caption{Comparison between attention distribution and information flow on the critical token location task. A significant difference in the distributions of critical and irrelevant contexts is revealed. }
    \label{fig:pre1}
\end{figure}

\begin{figure}[t]
    \centering
    \begin{minipage}[t]{0.485\linewidth}
        \centering
        \includegraphics[width=\linewidth]{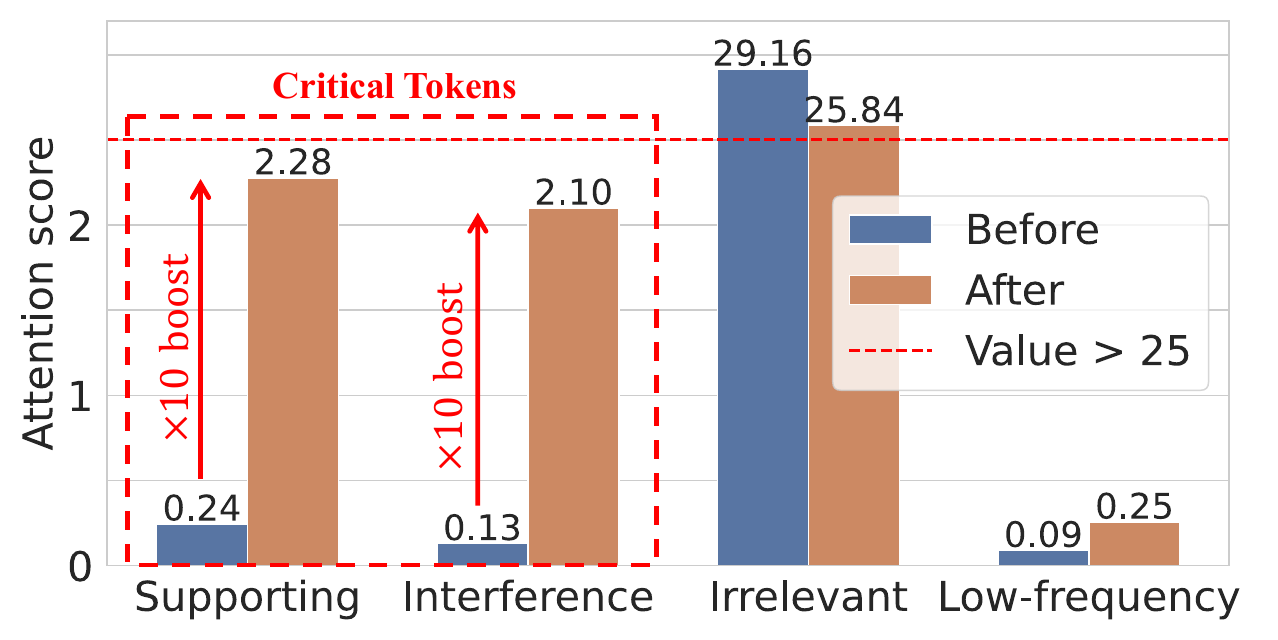}
        \caption{Attention distributions before and after manual context denoising. After context denoising, attention scores on critical tokens boost $\times 10$ times, and show a reduction on irrelevant tokens.}
        \label{fig:pre2_b}
    \end{minipage}
    \hfill
    \begin{minipage}[t]{0.485\linewidth}
        \centering
        \includegraphics[width=\linewidth]{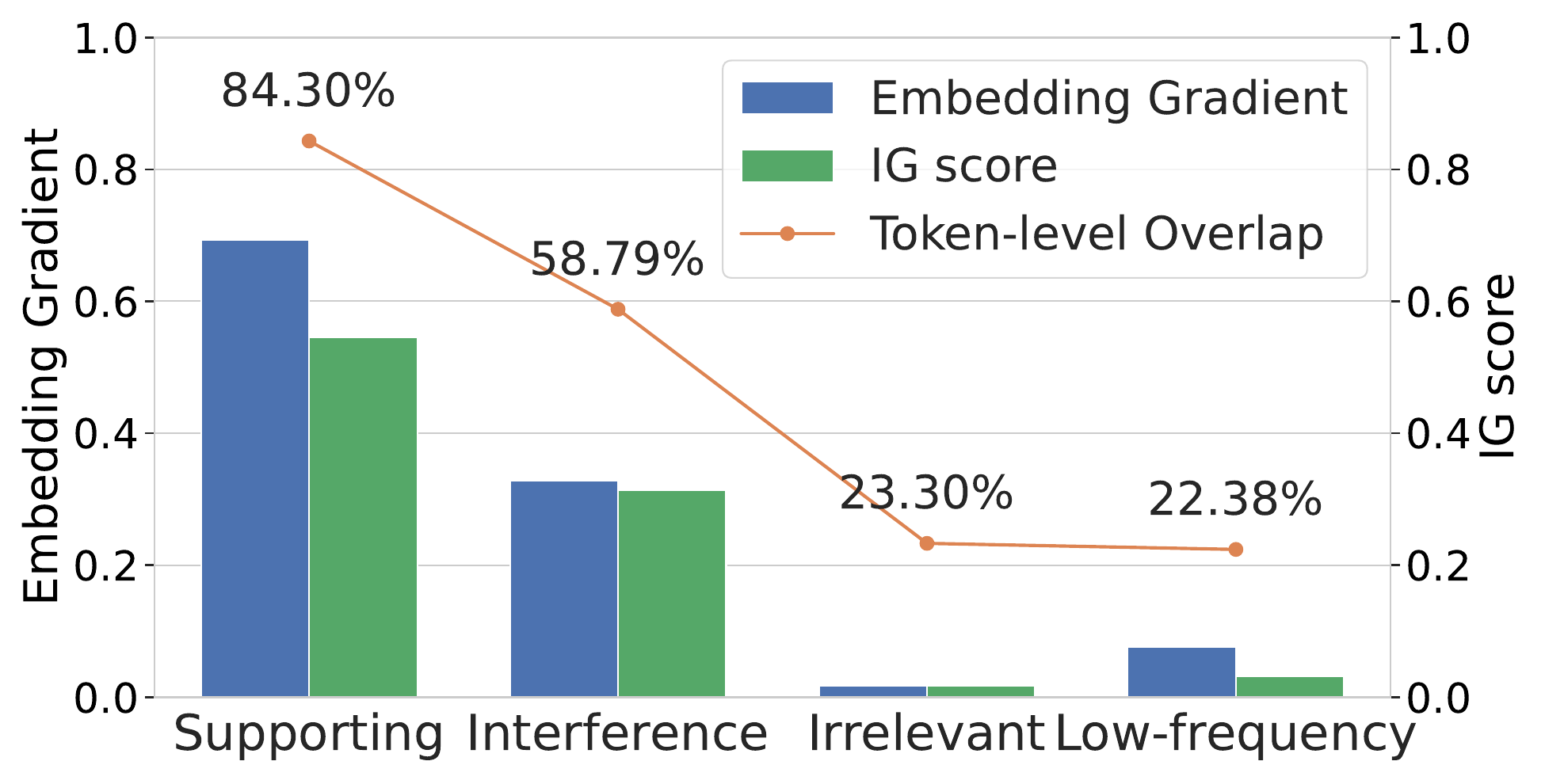}
        \caption{Relationship between attention IG score and L2-normalized embedding gradients on different types of tokens. It shows a proportional correlation.}
        \label{fig:pre2_a}
    \end{minipage}
\end{figure}

\paragraph{Information Flow Metric: IG score}
To discover the attention interaction among tokens, i.e., information flow~\citep{simonyan2013deep}, we employ the Integrated Gradient~(IG) technique~\citep{wang2023label}.
We define the IG score of $h$-th head in model's $l$-th layer on segment $\mathcal{T}_{r}$ below:
\begin{align}
    \mathrm{IG}_{h,l} = A_{h,l}^{T}\odot\mid \frac{\partial \mathcal{L}_\theta(Y|X)}{\partial A_{h,l}}\mid, ~~~~\mathrm{IG}^{(r)}_{h,l}=\frac{1}{|\mathcal{T}_{r}|}\sum\limits_{x_{i}\in\mathcal{T}_{r}}\sum\limits_{y_{j}\in Y}\mathrm{IG}_{h,l}[i, j],
    \label{eq:1}
\end{align}
where $\mathcal{L}_\theta(Y|X)$ is the model's prediction loss on $Y$, and $A_{h,l}$ denotes the attention matrix of the $h$-th head in the $l$-th layer.
The resulting IG score is a matrix, where each entry $\mathrm{IG}_{h,l}[i, j]$ represents the estimated bidirectional information flow between token $x_i$ and token $y_j$.
To assess the overall impact of $\mathcal{T}_{r}$ to $Y$, we compute the total contribution of tokens in $\mathcal{T}_{r}$ to the final prediction $Y$, i.e., $\mathrm{IG}^{(r)}_{h,l}$ and average across all attention heads and layers as the final score, i.e., $\mathrm{IG}^{(r)}$.
A higher IG score $\mathrm{IG}^{(r)}$ indicates a larger contribution from $\mathcal{T}_{r}$ to $Y$. 
Details are shown in Appendix~\ref{appdix:design_of_ig_score}.

\paragraph{Observation}
For a clear comparison, we normalize the computed FR and IG scores, and plot them in  Figure~\ref{fig:pre1}.
We find that the IG score detects significantly less noise~(irrelevant documents and low-frequency tokens) compared to the FR score on critical token detection. 
Specifically, as shown in Figure~\ref{fig:pre1_attn}, attention-based metrics reflect the distribution of tokens that the model focuses on during the generation process.
When the model generates correct responses, its attention focuses more on supporting facts; when the model generates wrong responses, its attention focuses more on interference tokens. 
\emph{Yet, in both cases, the FR score indicates that the model significantly focuses on irrelevant tokens.}
As for the IG score shown in Figure~\ref{fig:pre1_infor}, regardless of whether the response is correct or not, \emph{the IG score for critical tokens is significantly higher than that for irrelevant tokens}.
% Thus, we can effectively identify the critical tokens by leveraging the IG score and establish a connection between critical tokens and generation through subsequent training.

\subsection{Effect of Manual Context Noise Restraint}
\label{pre:sub2}
Considering that directly suppressing context noise in attention is very challenging, we aim to restrain the noise from the input perspective.
We first identify irrelevant tokens by computing the IG score on each token and treating the token with the IG score lower than a threshold as the noisy token.
Then, we manually suppressed their influence by subtracting the corresponding gradients from their input embeddings.
This is motivated by the fact that the model has largely converged on these noisy tokens, resulting in their gradients exhibiting low sensitivity.
As shown in Figure~\ref{fig:pre2_b}, we observe that after manual context denoising, the attention scores on critical tokens increase nearly $\times 10$ times, while the attention scores on irrelevant contextual tokens exhibit a slight decrease.
It is worth noting that this operation can be roughly analogized to \textit{denoising in the digital signal processing field}~\citep{kopsinis2009development}, as it reduces noise in the input sequence, allowing the model to focus more effectively on the under-fitting critical tokens.
\begin{figure}[t]
    \centering
    \includegraphics[width=\textwidth]{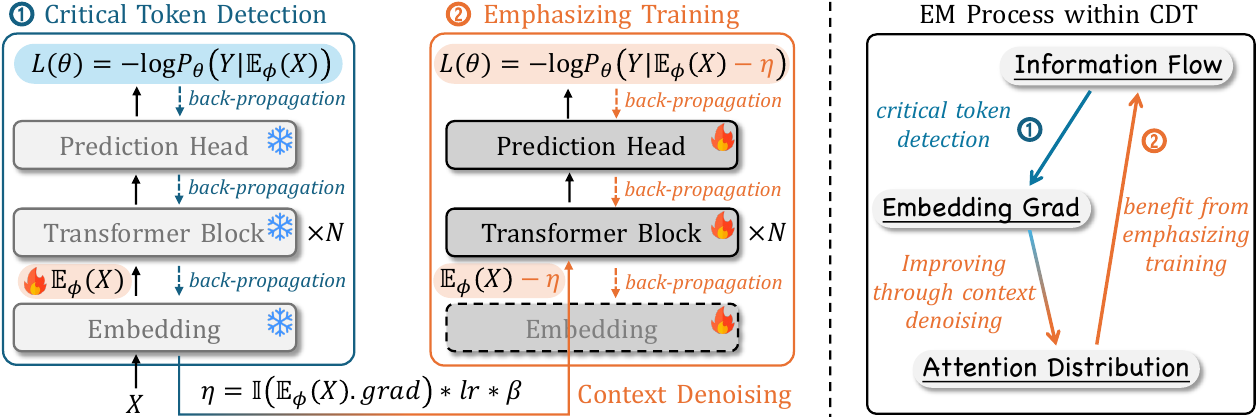}
    \caption{Our proposed CDT~(context denoising training) method. It consists of two steps: (1) detecting critical tokens within the long context, and (2) utilizing the denoised context for further emphasizing training. Notably, CDT can be understood as an \textit{Expectation Maximization~(EM)} process, where the model detects noise based on information flow and improves the training by diminishing the noise, thereby enhancing the information flow.}
    \label{fig:cdt_method}
\end{figure}

\section{Context Denoising Training}
\label{sec:CDO}
Based on the above observation, we propose a simple yet effective Context Denoising Training~(\textbf{CDT}) strategy.
Building upon the conventional language modeling objective, i.e., cross-entropy loss, CDT explicitly suppresses context noise during training to strengthen the model's attention on critical tokens and help establish a better connection between critical tokens and the final prediction.
It involves two key steps: (1) Critical Token Detection and (2) Emphasizing Training.
\vspace{-0.4em}
\subsection{Critical Token Detection}
\label{subsec:critical_token_detect}
Intuitively, we can first apply IG score to detect the critical tokens for the subsequent training.
However, computing the IG score in long-context scenario is highly GPU memory-intensive, as it requires storing full attention gradients and weights from every model layer across the entire sequence.
\textbf{Even with 8$\times$92GB GPUs~(H20), the maximum computable sequence length for the IG score is limited to 12K}, making it infeasible to generalize to a longer sequence.
Therefore, we designed a simple alternative implementation, which approximates the IG score with token embedding gradients\footnote{We choose token embeddings for 3 reasons: (1) they are easily accessible, (2) the embedding gradients are directly associated with tokens, and (3) they require much less GPU memory compared to attention gradients.}.
We derive a proportional relationship between the token embedding gradient and the IG score, and visualize the results in Figure~\ref{fig:pre2_a}. 
A detailed derivation is provided in Appendix~\ref{appedix:proof}.

As shown in Figure~\ref{fig:cdt_method}, given the input sequence $X = \{x_{i}\}_{i=1}^{n}$, label $Y$, and the model $f_{\theta}$, we first freeze the model parameters, keeping only the gradients of the input token embeddings $E_{\phi}(X)$, where $\phi\subset \theta$. 
We then obtain the gradient of each token embedding through the computation of the cross-entropy~(CE) loss followed by a loss back-propagation.
To identify the critical tokens, we employ an identifier $\mathbb{I}(\cdot)$ to detect tokens with large gradients, i.e., critical tokens, in the sequence.
Specifically, we define the calculation of the significance of each token as comparing its L2-normalized embedding gradient against the average of the computed gradients of all tokens, which can be written as:
\begin{equation}
\begin{aligned}
&\mathbb{I}(x_i) = 
\left\{
\begin{aligned}
    & 1, ~~\mathrm{if}~ ||\nabla_{E_{\phi}(x_i)}\mathcal{L}_{\mathrm{CE}}(x_i)||_{2} < t\\
    & 0, ~~\mathrm{if}~ ||\nabla_{E_{\phi}(x_i)}\mathcal{L}_{\mathrm{CE}}(x_i)||_{2} \geq t
\end{aligned}
\right.
&,~~t = \frac{1}{n}\sum_{i=1}^{n}||\nabla_{E_{\phi}(x_i)}\mathcal{L}_{\mathrm{CE}}(x_i)||_{2},
\end{aligned}
\end{equation}
\vspace{-0.4em}
where $\mathbb{I}(x_i)=1$ means $x_i$ is the irrelevant token~(noise); otherwise, it is critical token.

\subsection{Emphasizing Training}
\label{subsec:emphasize_train}
To suppress the context noise, we leverage the computed gradients to manipulate the irrelevant token embeddings, leaving critical tokens unchanged. The denoised token embedding can be denoised as: 
\begin{equation}
    E_{\phi}(x_i)^{\prime} = E_{\phi}(x_i) - \mathbb{I}(x_i)\nabla_{E_{\phi}(x_i)}\times lr\times\beta,
\label{equ:denoise}
\end{equation}
where $lr$ is the learning rate and $\beta$ is the hyper-parameter controlling the denoising level.
Then, we unfreeze the model and use the denoised token embeddings as the model input for further training, which we refer to as Emphasizing Training.
The loss function can be formulated as:
\begin{equation}
    \mathcal{L}_{CDT}(X, Y) = \mathcal{L}_{CE}\left(f_\theta\left(E_{\phi}(X)^{\prime}\right), Y \right).
\end{equation}

\paragraph{Remark}
Notably, the above process is conducted online during training rather than pre-computed offline. 
As shown in Figure~\ref{fig:cdt_method}, although this introduces additional computational overhead, CDT bootstraps the model’s long-context capabilities in an \textit{Expectation-Maximization (EM) manner}: the model first identifies the critical tokens based on information flow and improves the training by diminishing the noise, thereby ultimately enhancing the information flow.
In \S~\ref{subsec:step_performance}, we will demonstrate that, by training with CDT, the model can continuously enhance its performance compared to conventional training objectives during the post-training stage.

\section{Experiment}
\label{sec:experiment}

\subsection{Experimental Setups}
\label{subsec:settings_exp}

\paragraph{Evaluation}
We evaluate models on 4 different types of long-context tasks, including real-world tasks~(LongBench-E~\citep{bai2024longbench}, language modeling task~(LongPPL~\citep{fang2024wrong}), long-form reasoning task~(BABILong~\citep{kuratov2024babilong}), and synthetic tasks~(RULER~\citep{hsieh2024ruler}).
We compare CDT against existing widely-used methods on two types of models: (1) short-context models~(SCMs) that require context window scaling; (2) long-context models~(LCMs) that require long-context alignment.
In our main experiments, we select Llama-3-8B-Base model as the SCM, of which context window size is scaled $\times 8$ times~(64K).
For LCMs, we select Llama-3.1-8B-Base and Llama-3.1-8B-Instruct models.
We provide more evaluation and baseline details in Appendix~\ref{appdix:exp_details}, and show more evaluation results, such as generalizing CDT to more models,e.g., Qwen-series~\citep{yang2024qwen2,yang2025qwen3}), in Appendix~\ref{appdix:more_eval_res}.
We evaluate against current strong LCMs, as well as diverse long-context enhancement methods across training and inference paradigms — including token-wise reweighting~(LongCE~\citep{fang2024wrong}), KV-cache prefilling~\citep{laiflexprefill}, SFT~\citep{chen2024essential}, and RL-based optimization~\citep{tang2024logo}.

\paragraph{Training and Datasets}
For context window scaling training on SCM and post-training on LCM-Base, we apply PG-19~\citep{raecompressive2019} as the training data. 
For each training sample, we organize it into 64K tokens and collect 10,000 training samples.
For long-context alignment on LCM-Instruct, we utilize data sampled from LongMiT~\citep{chen2024essential} and LongAlpaca~\citep{long-alpaca}, covering 8,000 samples with context lengths ranging from 16K to 128K.
Based on the analysis experiment~(Section~\ref{subsec:impact_context_denoising}), we set $\beta=5$ in Equation~\ref{equ:denoise} for Llama-3.1 and Llama-3 models in the main experiments.
More dataset processing and implementation details are shown in Appendix~\ref{appdix:exp_details}

\begin{table*}[t]
\centering
\caption{Evaluation results on LongBench-E benchmark. To ensure fairness, we place existing works that do not use the same training data with us in the top group. 
Our method is implemeted under three settings: context-window scaling~(CWS), language modeling~(LM), and SFT.
% More evaluation and baseline details are shown in Appendix~\ref{appdix:exp_details}, and more evaluation results, such as generalizing CDT to more models,e.g., Qwen-series~\citep{yang2024qwen2,yang2025qwen3}), are listed in Appendix~\ref{appdix:more_eval_res}.
}
\resizebox{\textwidth}{!}{
\begin{tabular}{l c c c c c c c}
\toprule
\textbf{Models} & \textbf{Type} & \textbf{S-Doc QA} & \textbf{M-Doc QA} & \textbf{Summ} & \textbf{Few-shot} & \textbf{Code} & \textbf{Avg.}  \\
\midrule
ProLong-512K-Instruct~\citep{gao2024prolong} & SFT & 40.07 & 41.24 & 28.27 & 64.21 & 63.08 & 47.37  \\
NExtLong-512K-Instruct~\citep{gao2025nextlong} & SFT & 43.47 & 43.21 & 29.88 & 60.87 & 44.35 & 44.35 \\
Llama-3.1-8B-SEALONG~\citep{li2024large} & DPO & 49.45 & 44.69 & \bf 30.96 & 61.54 & 57.85 & 48.90 \\
GPT-4o~(version: 2024-11-20) & - & \bf 51.43 & \bf 60.89 & 14.78 & \bf 66.37 & \bf 61.25 & \bf 51.00 \\
\midrule
\multicolumn{8}{l}{\underline{\textbf{Results on Short-context Model}} (\textit{all SCMs share the same training data, $8\times$ context window scaling.})} \\
\midrule
Llama-3-8B-Base~(8K) & - & 25.20 & 21.52 & 20.18 & 32.67 & 27.92 & 25.50   \\
~~~+ YaRN~\citep{pengyarn} & - & 24.37 & 19.86 & 24.32 & 29.99 & 31.67 & 26.04  \\
~~~+ CE & CWS & \bf 25.29 &	21.49 &	20.36 &	32.69 &	27.76 &	34.62 \\
~~~+ LongCE~\citep{fang2024wrong} & CWS & 17.13 & 9.59 & 25.00 & 59.57 & 61.83 & 34.62 \\
~~~+ CDT~(ours) & CWS & 17.03 & \bf 24.87 &	\bf 26.61 &	\bf 61.89 &	\bf 66.14 &	\bf39.31   \\
\midrule
\multicolumn{8}{l}{\underline{\textbf{Results on Long-context Base Model}} (\textit{all LCMs share the same training data.})} \\
\midrule

Llama-3.1-8B-Base & - & 18.20 & \bf 13.19 & 21.13 & \bf 63.80 & 69.32 & 37.13 \\
~~~+ CE & LM & 17.10  & 10.82  & 26.38  & 62.85  & \bf 70.62  & 37.55   \\
~~~+ LongCE~\citep{fang2024wrong} & LM & 19.14 & 10.87 & 28.63 & 59.63 & 66.24 & 36.90   \\
~~~+ CDT~(ours) & LM & \bf 19.15	& 13.01	& \bf 29.23	& 63.63	& 69.44	& \bf 38.89 \\
\midrule
\multicolumn{8}{l}{\underline{\textbf{Results on Long-context Instruct Model}} (\textit{all LCMs use same source data with different formats.})} \\
\midrule

Llama-3.1-8B-Instruct & - & 48.58 &	45.19 &	30.30 & 61.73 &	57.26 &	48.61 \\
~~~+ Contriever~\citep{izacard2021contriever} & RAG & 42.63 &	45.55 & 32.48 & 62.15 & 41.85 & 44.93 \\
~~~+ FlexPrefill~\citep{laiflexprefill} & KV-Prefill & 47.02 & 45.55 & 27.37	& 60.97	& 55.97	& 47.38 \\
~~~+ X-Attention~\citep{xu2025xattention} & KV-Prefill & 48.32 & 45.60 & 26.93 & 61.83 & 56.39 & 47.81 \\
~~~+ SFT & SFT & 49.23 & 44.86 & 30.39 & 61.96 & 57.14 & 48.72 \\
~~~+ LOGO~\citep{tang2024logo} & DPO & 49.63 & 45.39 & \bf 30.44 &	62.39 &	57.19 &	49.01 \\
~~~+ CDT~(ours) & SFT & \bf 50.11 & \bf 46.04 & 30.34 & \bf 62.49 & \bf 65.64 & \bf 50.92 \\
\bottomrule
\end{tabular}}
\label{tab:longbench}
\end{table*}

\subsection{Results}
\paragraph{Real-world Long-context Understanding Tasks}
LongBench-E is a comprehensive benchmark suite encompassing 12 real-world datasets and various context lengths spread across 5 categories.
As shown in table~\ref{tab:longbench}, we observe that: \textbf{(1)} \emph{ CDT achieves the best performance among all the sub-tasks}. 
For SCMs, with the same training data, CDT achieved the best performance, outperforming a competitive counterpart~(LongCE) by nearly 4.7 points on average.
\textbf{(2)} For LCM-Base models, we find when post-training on the base model with language modeling training objective, \emph{CDT is the only method that ensures no significant performance drop across all subtasks}, and it even achieves some improvements. 
In contrast, using standard CE or LongCE objective leads to significant performance drops on some sub-tasks. 
For example, LongCE results in a nearly 4-point drop compared to the backbone model on the Few-shot subtask.
\textbf{(3)} As for the LCM-Instruct models~(the bottom group), we find that, due to its remarkable performance, \emph{existing post-training methods do not bring significant improvements}.
For instance, Llama-3.1-8B-SEALONG~(48.90) achieves only around slight 0.3-point average improvement compared to Llama-3.1-8B-Instruct~(49.61).
However, our CDT achieves an average improvement of more than 2 points compared to that of the backbone model across all tasks. We provide more analysis of results in Appendix~\ref{appdix:cdt_scale_more_models}.

\begin{table}[t]
    \centering
    \caption{Evaluation results on long synthetic tasks~(RULER), language modeling, and long-form reasoning~(BABILong). For RULER, we report the average scores across 13 sampled sub-tasks. To calculate LongPPL, we apply the Llama3-8B-Base model as the evaluator. For BABILong, we report the model reasoning capability from short context~(4K) to long context~(128K).}
    \resizebox{\linewidth}{!}{
    \begin{tabular}{l c c c| c | c c c c c c c}
    \toprule
    \multirow{2}{*}{\bf Models} & \multicolumn{3}{c}{\bf RULER} & \multicolumn{1}{|c}{\bf \makecell{Language \\ Modeling}} & \multicolumn{6}{|c}{\bf BABILong} \\
    \cmidrule{2-12} 
    & \bf 32K & \bf 64K & \bf 128K & \bf LongPPL & \bf 4K & \bf8K & \bf16K & \bf32K & \bf64K & \bf128K & \bf Avg. \\
    \midrule
    Llama-3-8B-Base & - & - & - &  > 100 & 33.40 & 26.60 & 4.80 & 0.00 & 0.20 & - & 13.00 \\
    ~~~+ YaRN & 39.58 & 31.46 & - & 3.55 & 35.20 & 29.80 & 24.40 & 20.20 & 17.60 & - &25.44  \\ 
    ~~~+ CE & 36.01 & 13.82	& - & 3.90	&  36.60 & 34.80 & 26.60 & 28.20 & 21.60 & - &29.56 \\
    ~~~+ LongCE & 84.02 & 71.50	& - & 3.55 & 36.00 & \bf 34.80 & 34.60 & \bf 32.60 & 29.40 & - &33.48  \\ 
    ~~~+ CDT~(ours) & \bf 84.76 & \bf 73.40	& - & \bf 3.04 & \bf 38.40 & 34.60 & \bf 34.80 & 31.40 & \bf 29.60 & - & \bf 33.76 \\
    \midrule                 
    Llama-3.1-8B-Base & 89.99 & 81.96 & 70.60 & 3.22 & 35.00 & 33.20 & 27.80 & 28.00 & 25.20 & 24.40 & 28.93\\
    ~~~+ CE & 86.59	& 80.87	& 70.44 & 3.28 & \bf 39.20 & 31.60 & 31.40 & 26.60 & 26.80 & 19.40 & 29.17\\
    ~~~+ LongCE & 87.65	& 81.79 & 70.79 & 3.24 & 37.80 & 33.40 & \bf 33.60 & \bf 32.60 & 27.60 & 24.60 & 31.60\\
    ~~~+ CDT~(ours) & \bf 90.36 &\bf 82.23 & \bf 74.12 & \bf 2.10 & 38.80 & \bf 36.60 & 33.20 & 29.40 & \bf 28.20 & \bf 28.20 & \bf 32.40 \\
    \midrule
    Llama-3.1-8B-Instruct & 92.49 &85.98 & 76.71 & 4.05 & 46.60 & 49.60 & 42.40 & 38.80 & 37.00 & 29.60 & 40.67\\
    ~~~+ SFT & 92.49 & 86.22 & 77.33 & 3.31 & 47.00 & 49.40 & \bf 43.60 & 41.20 & 37.40 & 30.40 & 41.50\\
    ~~~+ LOGO & 92.54 & 86.93 & 77.68 & 4.11 & 48.20 & 50.00 & 42.60 & 42.20 & 37.40 & 31.60 & 42.00\\
    ~~~+ CDT~(ours) & \bf 93.08 & \bf 88.01 & \bf 78.72 & \bf 2.36 & \bf 51.40 & \bf 51.20 & 41.60 & \bf 44.00 & \bf 38.60 & \bf 33.00 & \bf 43.30\\
    \bottomrule
    \end{tabular}}
    \label{tab:lm}
\end{table}

\paragraph{Long Synthetic Task and Language Modeling}
For the long synthetic task, we evaluate the model's performance under 32K, 64K, and 128K context lengths. 
We choose 13 sub-tasks from the RULER benchmark and report the average results.
For the language modeling task, we calculate LongPPL~\citep{fang2024wrong} on the GovReport dataset~\citep{huang-etal-2021-efficient}.
Notably, LongPPL can potentially reflect the model's ability to locate salient tokens in the long context.
More implementation and calculation details are illustrated in Appendix~\ref{appdix:evaluation_results}.
We show the evaluation results in Table~\ref{tab:lm},where our CDT method achieves the best model performance on the RULER benchmark from 32K to 128K settings.
Besides, in the language model task, CDT exhibits the lowest LongPPL, indicating the great potential of CDT to locate salient tokens.

\paragraph{Short-context \& Long-form Reasoning Tasks}
We evaluate the model's long-form reasoning capabilities, as well as its short-context capability, on BABILong, a synthetic task that requires models to reason through multiple supporting facts hidden in contexts of varying lengths~(from 4K to 64K).
As shown in Table~\ref{tab:lm}, our CDT achieves the highest overall score in each group. 
Besides, we observe that our CDT does not compromise the model's performance on short-context tasks. 
For instance, in the 4K and 8K lengths, CDT achieves either the best or comparable results compared to other methods and backbone models.
\section{Ablation Study}
\label{sec:analysis}
In this section, we compare the accuracy of salient token detection of CDT with other detection methods in \S\ref{subsec:analysis_detect}.  
Then, we show the impact of context denoising on the training process in \S\ref{subsec:impact_context_denoising}.
Finally, we elaborate on the training budget of our CDT method in \S\ref{subsec:step_performance}.
Notably, to help better understand the effectiveness of our CDT method, we also analyze the attention map patterns to reveal how CDT influences the model's attention distribution in Appendix~\ref{appdix:analysis_attn_map}.

\subsection{Comparison of Critical Token Detection}
\label{subsec:analysis_detect}

\begin{figure}[t]
    \centering
    \begin{minipage}[t]{0.48\linewidth}
        \centering
        \includegraphics[width=\linewidth]{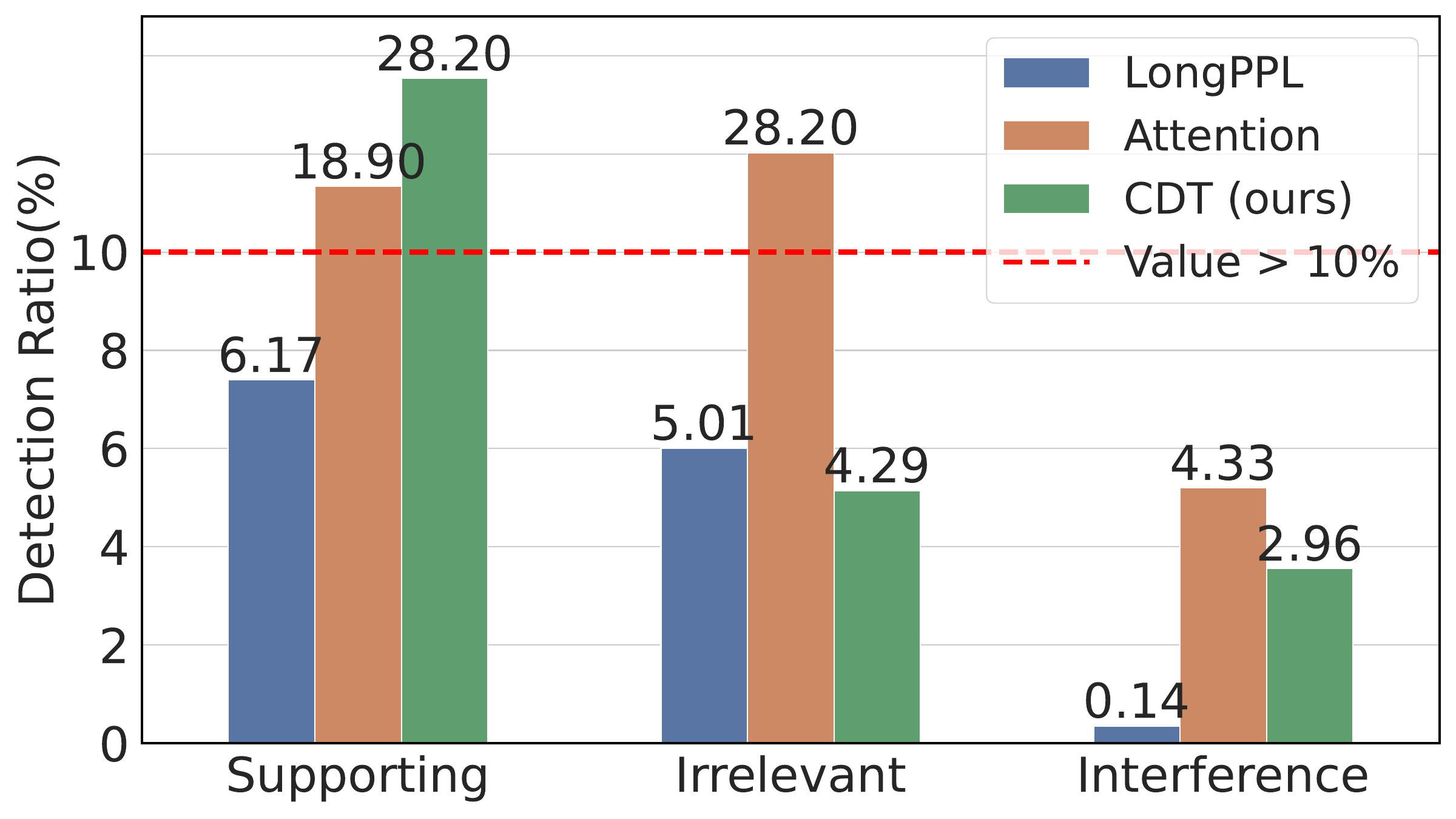}
        \caption{Comparison of critical token detection capability among different methods on our synthetic task. CDT achieves best performance.}
        \label{fig:detect}
    \end{minipage}
    \hfill
    \begin{minipage}[t]{0.48\linewidth}
        \centering
        \includegraphics[width=\linewidth]{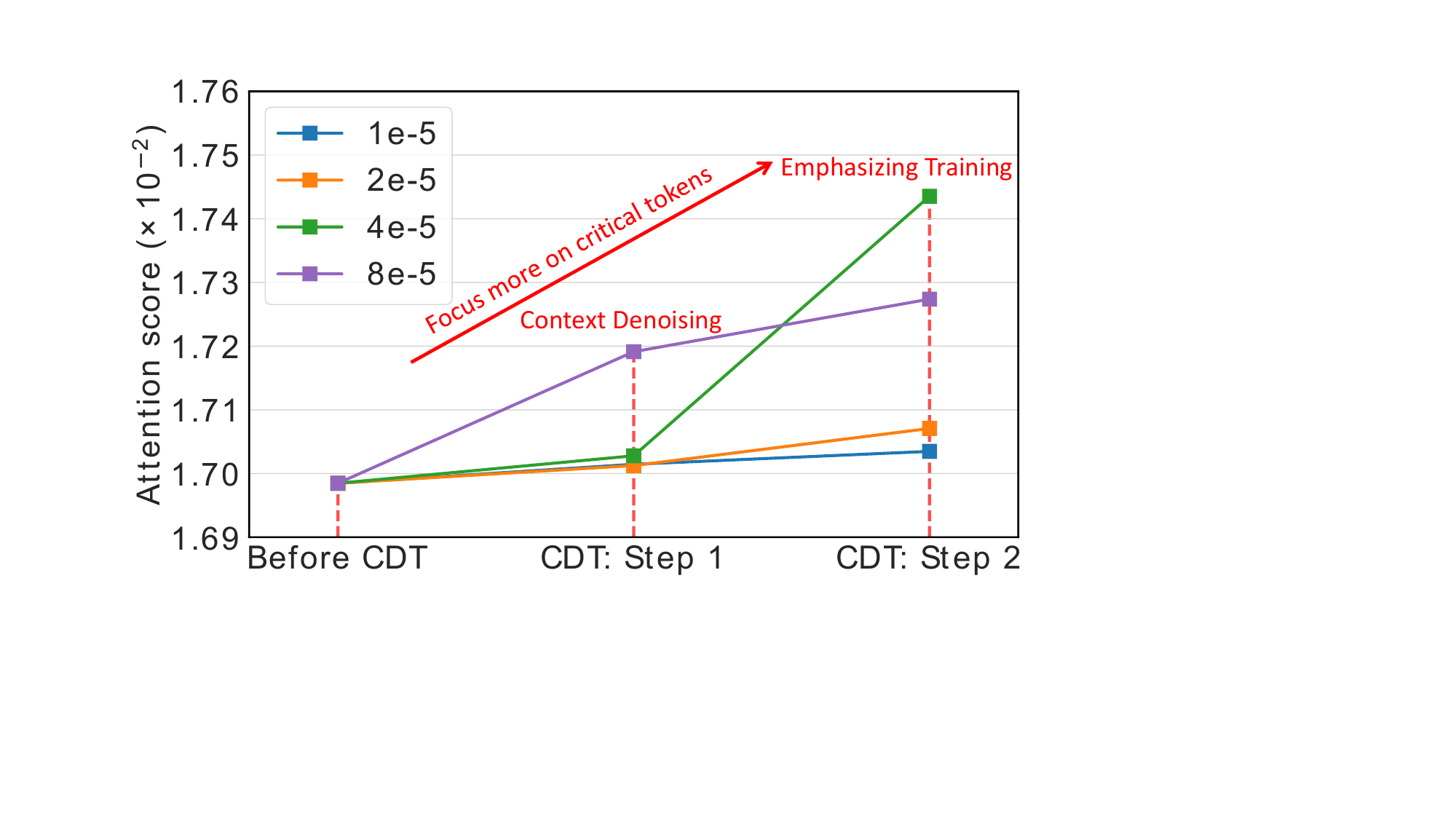}
        \caption{Impact of context denoising and comparison of the effect of learning rate on attention scores assigned to critical tokens in CDT.}
        \label{fig:train_analysis}
    \end{minipage}
    \vspace{-1em}
\end{figure}

We compare three different detection methods, including LongPPL, attention-based detection, and our CDT, on our synthetic task~(Figure~\ref{fig:pre_task_format}).
For attention-based and our CDT methods, we treat the tokens with the top-30 highest attention scores and L2 normalized gradient of embedding as the detected tokens.
As shown in Figure~\ref{fig:detect}, we can observe that the attention-based method can detect a high proportion of supporting tokens and interference tokens, but it also detects a large number of irrelevant tokens. 
On the other hand, while LongPPL can effectively suppress the detection of irrelevant tokens, it struggles to locate supporting tokens. 
Our CDT method not only identifies the largest number of critical tokens but also effectively suppresses the detection of irrelevant tokens.

\subsection{Impact of Context Denoising Strength}
\label{subsec:impact_context_denoising}
We visualize the changes in attention scores allocated to critical tokens during the CDT training process under different learning rates and the same $\beta=1$ settings.  
As shown in Figure~\ref{fig:train_analysis}, we observe that the attention scores on critical tokens have already increased significantly after the context denoising step. 
Furthermore, after the Emphasizing Training stage, there is an additional improvement. 
Additionally, we observe that a larger learning rate results in more pronounced improvements, further enhancing context denoising. 
However, a saturation point exists~(e.g., at 8e-5), beyond which the benefits plateau. 
Based on this observation, we adopt a learning rate $lr$ of 1e-5 and set $\beta=5$ in our main experiments, where $lr\times \beta=5e-5$.
We also recommend viewing the attention map provided in Appendix~\ref{appdix:analysis_attn_map}, which shows that CDT enables the model to focus more on key information within long context, without substantially changing the original attention distribution.

\begin{figure}[t]
    \centering
    \begin{minipage}[t]{0.48\linewidth}
        \centering
        \includegraphics[width=\linewidth]{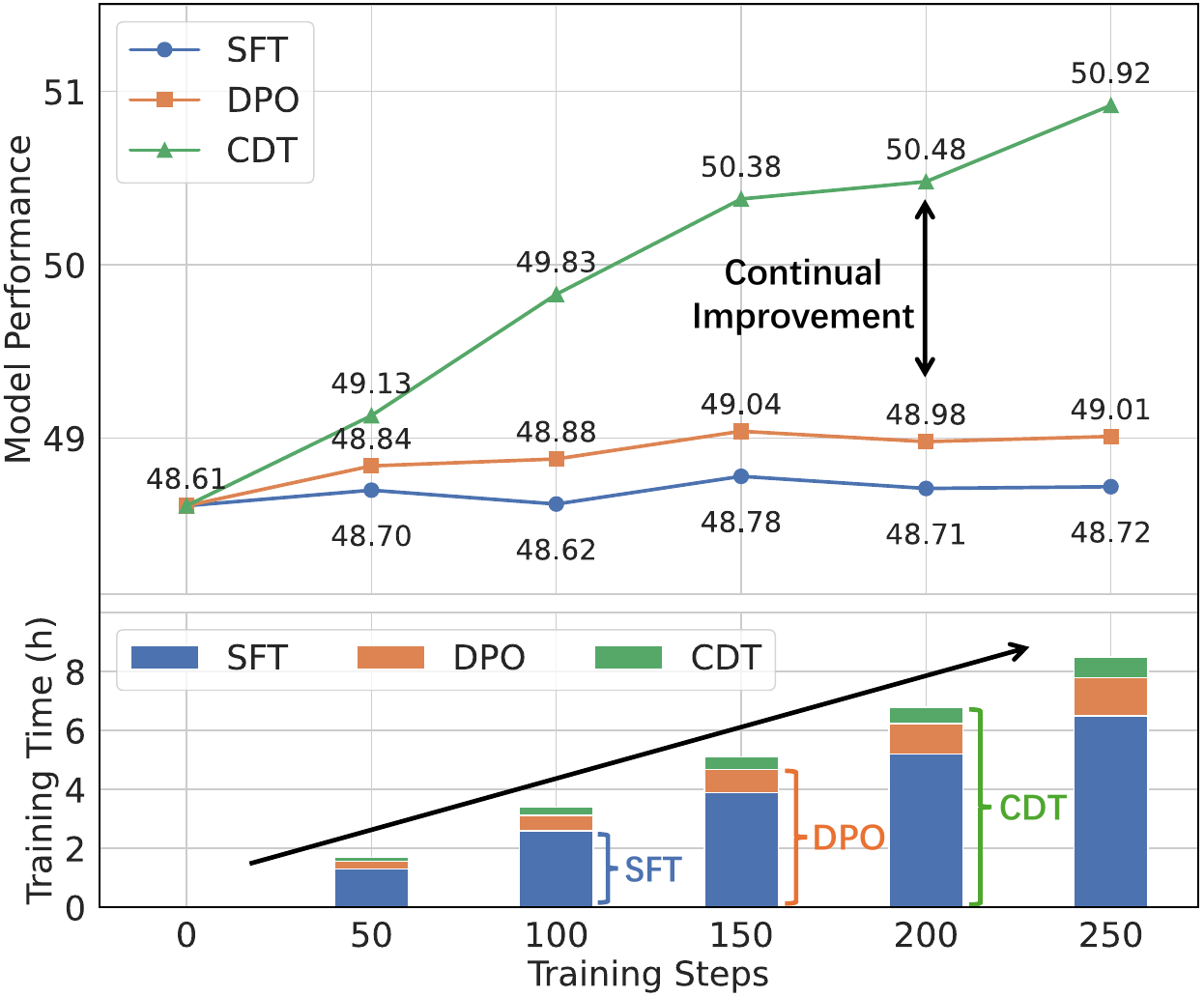}
        \caption{The performance improvement and training duration for every interval of 50 steps. With only a modest cost in training time, CDT significantly boosts the performance of LCM.}
        \label{fig:efficiency}  
    \end{minipage}
    \hfill
    \begin{minipage}[t]{0.48\linewidth}
        \centering
        \includegraphics[width=\linewidth]{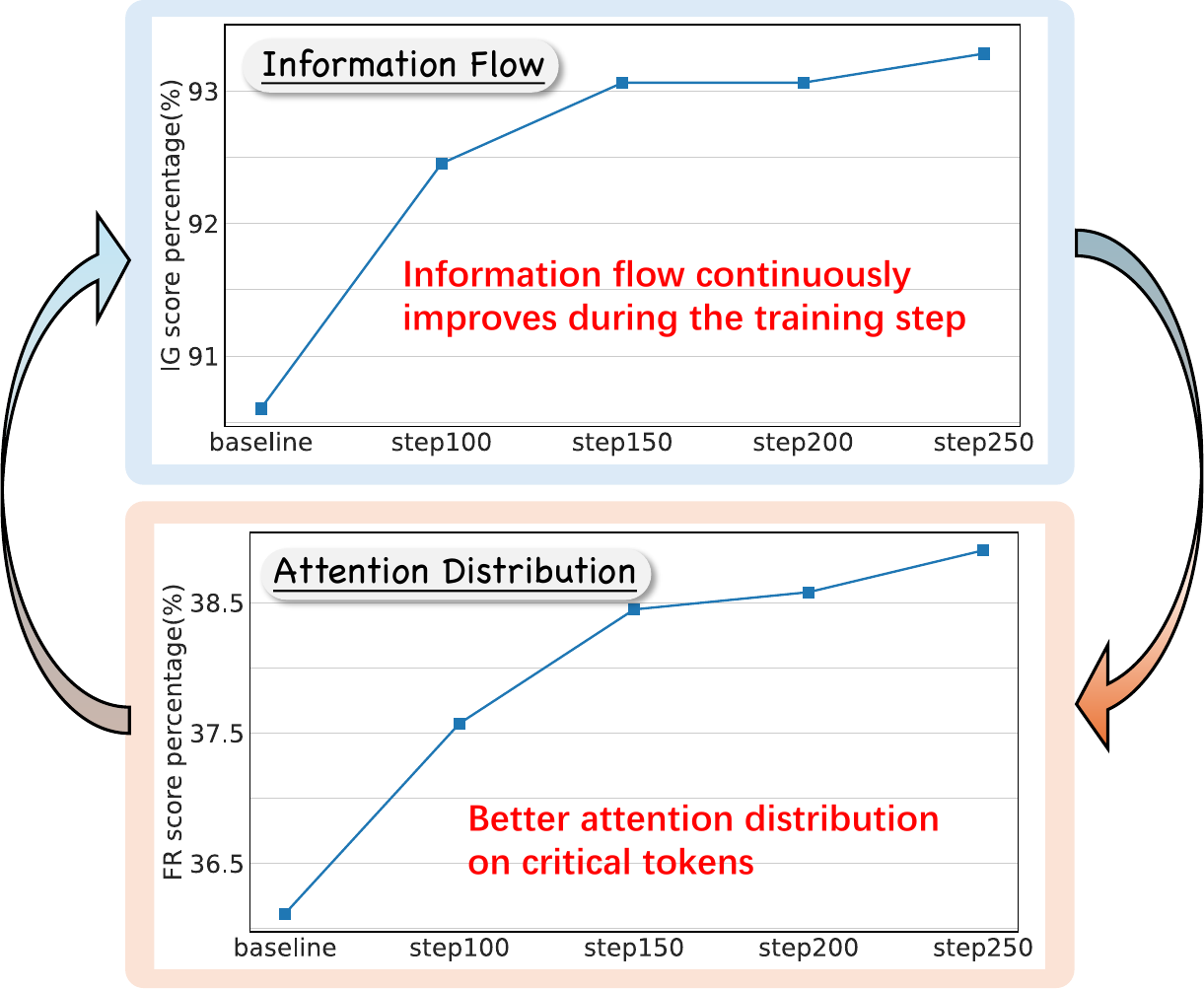}
        \caption{Illustration of \textit{EM} process of our CDT method, where both the information flow and attention distribution progressively improve within the training steps.}
        \label{fig:em_process_details}
    \end{minipage}
\end{figure}

\subsection{Training Budgets and \textit{EM} Process}
\label{subsec:step_performance}
Compared to conventional long-context training, which performs one forward and one backward pass to update all parameters, CDT introduces an additional noise detection step.
Critically, in long-context training, backward passes are typically 2–3$\times$ slower than forward passes due to activation recomputation~\citep{megatron-lm}.
Yet CDT adds merely one lightweight backward~(where the vast majority of model parameters are frozen) and one extra forward, resulting in minimal wall-clock overhead relative to standard training.
We compare CDT with SFT~(single Forward-Backward) and DPO~(one batch contains pairwise samples) methods.
As shown in Figure~\ref{fig:efficiency}, we observe that although CDT brings additional cost, i.e., approximately 0.5 hours in 8$\times$A100 GPUs for every 50 steps compared with SFT, it consistently and largely improves the model performance within the 250 training steps.
With the same training steps, DPO only yields marginal improvements, while SFT even demonstrates a decline in performance.
We provide the total training duration in Appendix~\ref{appdix:exp_details}. 
Such a great improvement can be largely attributed to the \textit{EM} process shown in Figure~\ref{fig:em_process_details}.
Notably, our approach exhibits a convergence boundary after approximately 250 steps.

\section{Conclusion}
\label{sec:conclusion}
Prior studies suggest that long-context models typically follow a \textit{retrieval-then-generation} paradigm, where the ``retrieval context'' may be overwhelmed by excessive irrelevant tokens. 
To address this issue, we present a fine-grained analysis of contextual noise in long-context inputs. 
We introduce a novel metric, the IG score, to effectively identify critical tokens, and observe that reducing contextual noise enables models to focus more precisely on critical tokens. 
Building on these insights, we propose Context Denoising Training~(CDT), a training strategy designed to both enhance the model's attention to critical tokens and strengthen the association between salient tokens and the model prediction.
Experiments across 4 task types~(including both short and long context length) and different models demonstrate the superiority of our method.
With CDT, an open-source 8B model can even achieve comparable performance with GPT-4o on real-world long-context tasks.

\section*{Ethics Statement}
\addcontentsline{toc}{section}{Ethics Statement}
We confirm that this work adheres to ethical research practices. All data and LLMs used are publicly available~(including API format) and properly cited. No human subjects were involved. 
The Use of LLM statement is illustrated in Appendix~\ref{appdix:use_of_llm}.

\section*{Reproducibility Statement}
All experimental settings, hyperparameters, and evaluation protocols are detailed in Section~\ref{subsec:settings_exp} and Appendix~\ref{appdix:exp_details}. Code, model checkpoints, and preliminary synthesis testing data will be released upon publication. Experiments are conducted on 8$\times$A100 GPUs with PyTorch, HuggingFace Transformers~\citep{wolf-etal-2020-transformers}, Deepspeed~\citep{rajbhandari2020zero} and LOOM-Scope~\citep{tang2025loom}.

\bibliography{iclr2026_conference}

\begin{thebibliography}{73}
\providecommand{\natexlab}[1]{#1}
\providecommand{\url}[1]{\texttt{#1}}
\expandafter\ifx\csname urlstyle\endcsname\relax
  \providecommand{\doi}[1]{doi: #1}\else
  \providecommand{\doi}{doi: \begingroup \urlstyle{rm}\Url}\fi

\bibitem[An et~al.(2024{\natexlab{a}})An, Zhang, Zhong, Li, Gong, Luo, Xu, and Kong]{an2024does}
Chenxin An, Jun Zhang, Ming Zhong, Lei Li, Shansan Gong, Yao Luo, Jingjing Xu, and Lingpeng Kong.
\newblock Why does the effective context length of llms fall short?
\newblock \emph{arXiv preprint arXiv:2410.18745}, 2024{\natexlab{a}}.

\bibitem[An et~al.(2024{\natexlab{b}})An, Ma, Lin, Zheng, and Lou]{an2024make}
Shengnan An, Zexiong Ma, Zeqi Lin, Nanning Zheng, and Jian-Guang Lou.
\newblock Make your llm fully utilize the context.
\newblock \emph{arXiv preprint arXiv:2404.16811}, 2024{\natexlab{b}}.

\bibitem[Bai et~al.(2024{\natexlab{a}})Bai, Lv, Zhang, He, Qi, Hou, Tang, Dong, and Li]{bai2024longalign}
Yushi Bai, Xin Lv, Jiajie Zhang, Yuze He, Ji~Qi, Lei Hou, Jie Tang, Yuxiao Dong, and Juanzi Li.
\newblock {L}ong{A}lign: A recipe for long context alignment of large language models.
\newblock In \emph{Findings of the Association for Computational Linguistics: EMNLP 2024}, pp.\  1376--1395, Miami, Florida, USA, November 2024{\natexlab{a}}. Association for Computational Linguistics.
\newblock \doi{10.18653/v1/2024.findings-emnlp.74}.
\newblock URL \url{https://aclanthology.org/2024.findings-emnlp.74}.

\bibitem[Bai et~al.(2024{\natexlab{b}})Bai, Lv, Zhang, Lyu, Tang, Huang, Du, Liu, Zeng, Hou, Dong, Tang, and Li]{bai2024longbench}
Yushi Bai, Xin Lv, Jiajie Zhang, Hongchang Lyu, Jiankai Tang, Zhidian Huang, Zhengxiao Du, Xiao Liu, Aohan Zeng, Lei Hou, Yuxiao Dong, Jie Tang, and Juanzi Li.
\newblock {L}ong{B}ench: A bilingual, multitask benchmark for long context understanding.
\newblock In \emph{Proceedings of the 62nd Annual Meeting of the Association for Computational Linguistics (Volume 1: Long Papers)}, pp.\  3119--3137, Bangkok, Thailand, August 2024{\natexlab{b}}. Association for Computational Linguistics.
\newblock \doi{10.18653/v1/2024.acl-long.172}.
\newblock URL \url{https://aclanthology.org/2024.acl-long.172}.

\bibitem[Bai et~al.(2024{\natexlab{c}})Bai, Tu, Zhang, Peng, Wang, Lv, Cao, Xu, Hou, Dong, Tang, and Li]{bai2024longbench2}
Yushi Bai, Shangqing Tu, Jiajie Zhang, Hao Peng, Xiaozhi Wang, Xin Lv, Shulin Cao, Jiazheng Xu, Lei Hou, Yuxiao Dong, Jie Tang, and Juanzi Li.
\newblock Longbench v2: Towards deeper understanding and reasoning on realistic long-context multitasks.
\newblock \emph{arXiv preprint arXiv:2412.15204}, 2024{\natexlab{c}}.

\bibitem[Bertsch et~al.(2024)Bertsch, Alon, Neubig, and Gormley]{bertsch2024unlimiformer}
Amanda Bertsch, Uri Alon, Graham Neubig, and Matthew Gormley.
\newblock Unlimiformer: Long-range transformers with unlimited length input.
\newblock \emph{Advances in Neural Information Processing Systems}, 36, 2024.

\bibitem[Chen et~al.(2023{\natexlab{a}})Chen, Wong, Chen, and Tian]{chen2023extending}
Shouyuan Chen, Sherman Wong, Liangjian Chen, and Yuandong Tian.
\newblock Extending context window of large language models via positional interpolation.
\newblock \emph{arXiv preprint arXiv:2306.15595}, 2023{\natexlab{a}}.

\bibitem[Chen et~al.(2023{\natexlab{b}})Chen, Qian, Tang, Lai, Liu, Han, and Jia]{chen2023longlora}
Yukang Chen, Shengju Qian, Haotian Tang, Xin Lai, Zhijian Liu, Song Han, and Jiaya Jia.
\newblock Longlora: Efficient fine-tuning of long-context large language models.
\newblock \emph{arXiv preprint arXiv:2309.12307}, 2023{\natexlab{b}}.

\bibitem[Chen et~al.(2023{\natexlab{c}})Chen, Yu, Qian, Tang, Lai, Liu, Han, and Jia]{long-alpaca}
Yukang Chen, Shaozuo Yu, Shengju Qian, Haotian Tang, Xin Lai, Zhijian Liu, Song Han, and Jiaya Jia.
\newblock Long alpaca: Long-context instruction-following models.
\newblock \url{https://github.com/dvlab-research/LongLoRA}, 2023{\natexlab{c}}.

\bibitem[Chen et~al.(2024)Chen, Chen, Qin, Guo, Lv, Zou, Che, Yan, Chen, and Lin]{chen2024essential}
Zhi Chen, Qiguang Chen, Libo Qin, Qipeng Guo, Haijun Lv, Yicheng Zou, Wanxiang Che, Hang Yan, Kai Chen, and Dahua Lin.
\newblock What are the essential factors in crafting effective long context multi-hop instruction datasets? insights and best practices.
\newblock \emph{arXiv preprint arXiv:2409.01893}, 2024.

\bibitem[Chevalier et~al.(2023)Chevalier, Wettig, Ajith, and Chen]{chevalier2023adapting}
Alexis Chevalier, Alexander Wettig, Anirudh Ajith, and Danqi Chen.
\newblock Adapting language models to compress contexts.
\newblock In \emph{Proceedings of the 2023 Conference on Empirical Methods in Natural Language Processing}, pp.\  3829--3846, 2023.

\bibitem[Ding et~al.(2024)Ding, Zhang, Zhang, Xu, Shang, Xu, Yang, and Yang]{ding2024longrope}
Yiran Ding, Li~Lyna Zhang, Chengruidong Zhang, Yuanyuan Xu, Ning Shang, Jiahang Xu, Fan Yang, and Mao Yang.
\newblock Longrope: Extending llm context window beyond 2 million tokens.
\newblock \emph{arXiv preprint arXiv:2402.13753}, 2024.

\bibitem[Fang et~al.(2024{\natexlab{a}})Fang, Miao, Srivastav, Liu, Zhang, Fang, Tsang, Nazari, Wang, Homayoun, et~al.]{fang2024large}
Chongzhou Fang, Ning Miao, Shaurya Srivastav, Jialin Liu, Ruoyu Zhang, Ruijie Fang, Ryan Tsang, Najmeh Nazari, Han Wang, Houman Homayoun, et~al.
\newblock Large language models for code analysis: Do $\{$LLMs$\}$ really do their job?
\newblock In \emph{33rd USENIX Security Symposium (USENIX Security 24)}, pp.\  829--846, 2024{\natexlab{a}}.

\bibitem[Fang et~al.(2024{\natexlab{b}})Fang, Wang, Liu, Zhang, Jegelka, Gao, Ding, and Wang]{fang2024wrong}
Lizhe Fang, Yifei Wang, Zhaoyang Liu, Chenheng Zhang, Stefanie Jegelka, Jinyang Gao, Bolin Ding, and Yisen Wang.
\newblock What is wrong with perplexity for long-context language modeling?
\newblock \emph{arXiv preprint arXiv:2410.23771}, 2024{\natexlab{b}}.

\bibitem[Fu et~al.(2024{\natexlab{a}})Fu, Panda, Niu, Yue, Hajishirzi, Kim, and Peng]{fudata}
Yao Fu, Rameswar Panda, Xinyao Niu, Xiang Yue, Hannaneh Hajishirzi, Yoon Kim, and Hao Peng.
\newblock Data engineering for scaling language models to 128k context.
\newblock In \emph{Forty-first International Conference on Machine Learning}, 2024{\natexlab{a}}.

\bibitem[Fu et~al.(2024{\natexlab{b}})Fu, Panda, Niu, Yue, Hajishirzi, Kim, and Peng]{fudata2024}
Yao Fu, Rameswar Panda, Xinyao Niu, Xiang Yue, Hannaneh Hajishirzi, Yoon Kim, and Hao Peng.
\newblock Data engineering for scaling language models to 128k context.
\newblock In \emph{Forty-first International Conference on Machine Learning}, 2024{\natexlab{b}}.

\bibitem[Gao et~al.(2024{\natexlab{a}})Gao, Wu, Fu, and Hu]{gao2024quest}
Chaochen Gao, Xing Wu, Qi~Fu, and Songlin Hu.
\newblock Quest: Query-centric data synthesis approach for long-context scaling of large language model.
\newblock \emph{arXiv preprint arXiv:2405.19846}, 2024{\natexlab{a}}.

\bibitem[Gao et~al.(2025)Gao, Wu, Lin, Zhang, and Hu]{gao2025nextlong}
Chaochen Gao, Xing Wu, Zijia Lin, Debing Zhang, and Songlin Hu.
\newblock Nextlong: Toward effective long-context training without long documents.
\newblock \emph{arXiv preprint arXiv:2501.12766}, 2025.

\bibitem[Gao et~al.(2024{\natexlab{b}})Gao, Wettig, Yen, and Chen]{gao2024prolong}
Tianyu Gao, Alexander Wettig, Howard Yen, and Danqi Chen.
\newblock How to train long-context language models (effectively).
\newblock \emph{arXiv preprint arXiv:2410.02660}, 2024{\natexlab{b}}.

\bibitem[Gao et~al.(2024{\natexlab{c}})Gao, Wettig, Yen, and Chen]{gao2024train}
Tianyu Gao, Alexander Wettig, Howard Yen, and Danqi Chen.
\newblock How to train long-context language models (effectively).
\newblock \emph{arXiv preprint arXiv:2410.02660}, 2024{\natexlab{c}}.

\bibitem[Ge et~al.(2025)Ge, Feng, Huang, Fu, Nie, Zuo, Lin, Cui, and Liu]{ge2025bytescale}
Hao Ge, Junda Feng, Qi~Huang, Fangcheng Fu, Xiaonan Nie, Lei Zuo, Haibin Lin, Bin Cui, and Xin Liu.
\newblock Bytescale: Efficient scaling of llm training with a 2048k context length on more than 12,000 gpus.
\newblock \emph{arXiv preprint arXiv:2502.21231}, 2025.

\bibitem[Gema et~al.(2024)Gema, Jin, Abdulaal, Diethe, Teare, Alex, Minervini, and Saseendran]{gema2024decore}
Aryo~Pradipta Gema, Chen Jin, Ahmed Abdulaal, Tom Diethe, Philip Teare, Beatrice Alex, Pasquale Minervini, and Amrutha Saseendran.
\newblock Decore: decoding by contrasting retrieval heads to mitigate hallucinations.
\newblock \emph{arXiv preprint arXiv:2410.18860}, 2024.

\bibitem[Guo et~al.(2025)Guo, Yang, Sun, Ding, Liu, and Lin]{guo2025learning}
Yiju Guo, Wenkai Yang, Zexu Sun, Ning Ding, Zhiyuan Liu, and Yankai Lin.
\newblock Learning to focus: Causal attention distillation via gradient-guided token pruning.
\newblock \emph{arXiv preprint arXiv:2506.07851}, 2025.

\bibitem[Helm et~al.(2025)Helm, Daheim, and Gurevych]{helm2025token}
Falko Helm, Nico Daheim, and Iryna Gurevych.
\newblock Token weighting for long-range language modeling.
\newblock \emph{arXiv preprint arXiv:2503.09202}, 2025.

\bibitem[Hsieh et~al.(2024)Hsieh, Sun, Kriman, Acharya, Rekesh, Jia, Zhang, and Ginsburg]{hsieh2024ruler}
Cheng-Ping Hsieh, Simeng Sun, Samuel Kriman, Shantanu Acharya, Dima Rekesh, Fei Jia, Yang Zhang, and Boris Ginsburg.
\newblock Ruler: What's the real context size of your long-context language models?
\newblock \emph{arXiv preprint arXiv:2404.06654}, 2024.

\bibitem[Hu et~al.(2024)Hu, Wu, Zhu, Xianyu, Wang, Zhang, and Cao]{hu2024openrlhf}
Jian Hu, Xibin Wu, Zilin Zhu, Xianyu, Weixun Wang, Dehao Zhang, and Yu~Cao.
\newblock Openrlhf: An easy-to-use, scalable and high-performance rlhf framework.
\newblock \emph{arXiv preprint arXiv:2405.11143}, 2024.

\bibitem[Huang et~al.(2021)Huang, Cao, Parulian, Ji, and Wang]{huang-etal-2021-efficient}
Luyang Huang, Shuyang Cao, Nikolaus Parulian, Heng Ji, and Lu~Wang.
\newblock Efficient attentions for long document summarization.
\newblock In \emph{Proceedings of the 2021 Conference of the North American Chapter of the Association for Computational Linguistics: Human Language Technologies}, pp.\  1419--1436, Online, June 2021. Association for Computational Linguistics.
\newblock \doi{10.18653/v1/2021.naacl-main.112}.
\newblock URL \url{https://aclanthology.org/2021.naacl-main.112}.

\bibitem[Izacard et~al.(2021)Izacard, Caron, Hosseini, Riedel, Bojanowski, Joulin, and Grave]{izacard2021contriever}
Gautier Izacard, Mathilde Caron, Lucas Hosseini, Sebastian Riedel, Piotr Bojanowski, Armand Joulin, and Edouard Grave.
\newblock Unsupervised dense information retrieval with contrastive learning, 2021.
\newblock URL \url{https://arxiv.org/abs/2112.09118}.

\bibitem[Jiang et~al.(2023)Jiang, Sablayrolles, Mensch, Bamford, Chaplot, Casas, Bressand, Lengyel, Lample, Saulnier, et~al.]{jiang2023mistral}
Albert~Q Jiang, Alexandre Sablayrolles, Arthur Mensch, Chris Bamford, Devendra~Singh Chaplot, Diego de~las Casas, Florian Bressand, Gianna Lengyel, Guillaume Lample, Lucile Saulnier, et~al.
\newblock Mistral 7b.
\newblock \emph{arXiv preprint arXiv:2310.06825}, 2023.

\bibitem[Kopsinis \& McLaughlin(2009)Kopsinis and McLaughlin]{kopsinis2009development}
Yannis Kopsinis and Stephen McLaughlin.
\newblock Development of emd-based denoising methods inspired by wavelet thresholding.
\newblock \emph{IEEE Transactions on signal Processing}, 57\penalty0 (4):\penalty0 1351--1362, 2009.

\bibitem[Kuratov et~al.(2024)Kuratov, Bulatov, Anokhin, Rodkin, Sorokin, Sorokin, and Burtsev]{kuratov2024babilong}
Yuri Kuratov, Aydar Bulatov, Petr Anokhin, Ivan Rodkin, Dmitry Sorokin, Artyom Sorokin, and Mikhail Burtsev.
\newblock Babilong: Testing the limits of llms with long context reasoning-in-a-haystack.
\newblock \emph{arXiv preprint arXiv:2406.10149}, 2024.

\bibitem[Lai et~al.(2025)Lai, Lu, Luo, Ma, and Zhou]{laiflexprefill}
Xunhao Lai, Jianqiao Lu, Yao Luo, Yiyuan Ma, and Xun Zhou.
\newblock Flexprefill: A context-aware sparse attention mechanism for efficient long-sequence inference.
\newblock In \emph{The Thirteenth International Conference on Learning Representations}, 2025.

\bibitem[Li et~al.(2024{\natexlab{a}})Li, Verga, Sen, Yang, Viswanathan, Lewis, Watanabe, and Su]{li2024alr}
Huayang Li, Pat Verga, Priyanka Sen, Bowen Yang, Vijay Viswanathan, Patrick Lewis, Taro Watanabe, and Yixuan Su.
\newblock Alr\^2: A retrieve-then-reason framework for long-context question answering.
\newblock \emph{arXiv preprint arXiv:2410.03227}, 2024{\natexlab{a}}.

\bibitem[Li et~al.(2024{\natexlab{b}})Li, Yang, Cheng, Liu, Yu, Yang, and Lam]{li2024large}
Siheng Li, Cheng Yang, Zesen Cheng, Lemao Liu, Mo~Yu, Yujiu Yang, and Wai Lam.
\newblock Large language models can self-improve in long-context reasoning.
\newblock \emph{arXiv preprint arXiv:2411.08147}, 2024{\natexlab{b}}.

\bibitem[Liu et~al.(2023)Liu, Zaharia, and Abbeel]{liu2023ring}
Hao Liu, Matei Zaharia, and Pieter Abbeel.
\newblock Ring attention with blockwise transformers for near-infinite context.
\newblock \emph{arXiv preprint arXiv:2310.01889}, 2023.

\bibitem[Liu et~al.(2024{\natexlab{a}})Liu, Bai, Zhang, Zhang, Zhang, Zhang, Wang, Que, Chen, Su, et~al.]{liu20242}
Jiaheng Liu, Zhiqi Bai, Yuanxing Zhang, Chenchen Zhang, Yu~Zhang, Ge~Zhang, Jiakai Wang, Haoran Que, Yukang Chen, Wenbo Su, et~al.
\newblock E\^{} 2-llm: Efficient and extreme length extension of large language models.
\newblock \emph{arXiv preprint arXiv:2401.06951}, 2024{\natexlab{a}}.

\bibitem[Liu et~al.(2024{\natexlab{b}})Liu, Lin, Hewitt, Paranjape, Bevilacqua, Petroni, and Liang]{liu2024lost}
Nelson~F Liu, Kevin Lin, John Hewitt, Ashwin Paranjape, Michele Bevilacqua, Fabio Petroni, and Percy Liang.
\newblock Lost in the middle: How language models use long contexts.
\newblock \emph{Transactions of the Association for Computational Linguistics}, 11:\penalty0 157--173, 2024{\natexlab{b}}.

\bibitem[Lu et~al.(2025)Lu, Jiang, Liu, Du, Jiang, Hong, Liu, He, Yuan, Wang, et~al.]{lu2025moba}
Enzhe Lu, Zhejun Jiang, Jingyuan Liu, Yulun Du, Tao Jiang, Chao Hong, Shaowei Liu, Weiran He, Enming Yuan, Yuzhi Wang, et~al.
\newblock Moba: Mixture of block attention for long-context llms.
\newblock \emph{arXiv preprint arXiv:2502.13189}, 2025.

\bibitem[Lu et~al.(2024)Lu, Yan, Yang, Chiu, Ren, Yuan, Zhao, Wu, and Rush]{lu2024controlled}
Yi~Lu, Jing~Nathan Yan, Songlin Yang, Justin~T Chiu, Siyu Ren, Fei Yuan, Wenting Zhao, Zhiyong Wu, and Alexander~M Rush.
\newblock A controlled study on long context extension and generalization in llms.
\newblock \emph{arXiv preprint arXiv:2409.12181}, 2024.

\bibitem[Luo et~al.(2025)Luo, Zhang, Yuan, Zhao, Yang, Gu, Wu, Chen, Qiao, Long, et~al.]{luo2025large}
Junyu Luo, Weizhi Zhang, Ye~Yuan, Yusheng Zhao, Junwei Yang, Yiyang Gu, Bohan Wu, Binqi Chen, Ziyue Qiao, Qingqing Long, et~al.
\newblock Large language model agent: A survey on methodology, applications and challenges.
\newblock \emph{arXiv preprint arXiv:2503.21460}, 2025.

\bibitem[Meta(2024)]{meta2024introducing}
AI~Meta.
\newblock Introducing llama 3.1: Our most capable models to date.
\newblock \emph{Meta AI Blog}, 12, 2024.

\bibitem[Meta(2025)]{meta2025llama}
AI~Meta.
\newblock The llama 4 herd: The beginning of a new era of natively multimodal ai innovation, april 2025, 2025.

\bibitem[MiniMax et~al.(2025)MiniMax, Li, Gong, Yang, Shan, Liu, Zhu, Zhang, Guo, Chen, Li, Jiao, Li, Zhang, Sun, Dong, Zhu, Zhuang, Song, Zhu, Han, Li, Xie, Xu, Yan, Zhang, Xiao, Kang, Han, Wang, Yu, Feng, Zheng, Chai, Xing, Ju, Chi, Zhang, Huang, Niu, Li, Zhao, Yang, Xu, Wang, Wang, Li, Leng, Shi, Yu, Li, Zhu, Huang, Liang, Sun, Sun, Cheng, Li, Song, Su, Han, Zhang, Hou, Min, Zou, Shen, Gong, Zhu, Zhou, Zhong, Hu, Fan, Yu, Yang, Li, Huang, Li, Huang, Xu, Mao, Li, Li, Tao, Ying, Cong, Qin, Fan, Yu, Jiang, and Wu]{minimax2025minimax01scalingfoundationmodels}
MiniMax, Aonian Li, Bangwei Gong, Bo~Yang, Boji Shan, Chang Liu, Cheng Zhu, Chunhao Zhang, Congchao Guo, Da~Chen, Dong Li, Enwei Jiao, Gengxin Li, Guojun Zhang, Haohai Sun, Houze Dong, Jiadai Zhu, Jiaqi Zhuang, Jiayuan Song, Jin Zhu, Jingtao Han, Jingyang Li, Junbin Xie, Junhao Xu, Junjie Yan, Kaishun Zhang, Kecheng Xiao, Kexi Kang, Le~Han, Leyang Wang, Lianfei Yu, Liheng Feng, Lin Zheng, Linbo Chai, Long Xing, Meizhi Ju, Mingyuan Chi, Mozhi Zhang, Peikai Huang, Pengcheng Niu, Pengfei Li, Pengyu Zhao, Qi~Yang, Qidi Xu, Qiexiang Wang, Qin Wang, Qiuhui Li, Ruitao Leng, Shengmin Shi, Shuqi Yu, Sichen Li, Songquan Zhu, Tao Huang, Tianrun Liang, Weigao Sun, Weixuan Sun, Weiyu Cheng, Wenkai Li, Xiangjun Song, Xiao Su, Xiaodong Han, Xinjie Zhang, Xinzhu Hou, Xu~Min, Xun Zou, Xuyang Shen, Yan Gong, Yingjie Zhu, Yipeng Zhou, Yiran Zhong, Yongyi Hu, Yuanxiang Fan, Yue Yu, Yufeng Yang, Yuhao Li, Yunan Huang, Yunji Li, Yunpeng Huang, Yunzhi Xu, Yuxin Mao, Zehan Li, Zekang Li, Zewei Tao, Zewen Ying, Zhaoyang Cong, Zhen
  Qin, Zhenhua Fan, Zhihang Yu, Zhuo Jiang, and Zijia Wu.
\newblock Minimax-01: Scaling foundation models with lightning attention, 2025.
\newblock URL \url{https://arxiv.org/abs/2501.08313}.

\bibitem[Peng et~al.(2023)Peng, Quesnelle, Fan, and Shippole]{pengyarn}
Bowen Peng, Jeffrey Quesnelle, Honglu Fan, and Enrico Shippole.
\newblock Yarn: Efficient context window extension of large language models.
\newblock In \emph{The Twelfth International Conference on Learning Representations}, 2023.

\bibitem[Qiu et~al.(2025{\natexlab{a}})Qiu, Embar, Zhang, Jaitly, Cohen, and Han]{qiu2025eliciting}
Yifu Qiu, Varun Embar, Yizhe Zhang, Navdeep Jaitly, Shay~B Cohen, and Benjamin Han.
\newblock Eliciting in-context retrieval and reasoning for long-context large language models.
\newblock \emph{arXiv preprint arXiv:2501.08248}, 2025{\natexlab{a}}.

\bibitem[Qiu et~al.(2025{\natexlab{b}})Qiu, Wang, Zheng, Huang, Wen, Yang, Men, Yu, Huang, Huang, et~al.]{qiu2025gated}
Zihan Qiu, Zekun Wang, Bo~Zheng, Zeyu Huang, Kaiyue Wen, Songlin Yang, Rui Men, Le~Yu, Fei Huang, Suozhi Huang, et~al.
\newblock Gated attention for large language models: Non-linearity, sparsity, and attention-sink-free.
\newblock \emph{arXiv preprint arXiv:2505.06708}, 2025{\natexlab{b}}.

\bibitem[Rae et~al.(2019)Rae, Potapenko, Jayakumar, Hillier, and Lillicrap]{raecompressive2019}
Jack~W Rae, Anna Potapenko, Siddhant~M Jayakumar, Chloe Hillier, and Timothy~P Lillicrap.
\newblock Compressive transformers for long-range sequence modelling.
\newblock \emph{arXiv preprint}, 2019.
\newblock URL \url{https://arxiv.org/abs/1911.05507}.

\bibitem[Rajbhandari et~al.(2020)Rajbhandari, Rasley, Ruwase, and He]{rajbhandari2020zero}
Samyam Rajbhandari, Jeff Rasley, Olatunji Ruwase, and Yuxiong He.
\newblock Zero: Memory optimizations toward training trillion parameter models.
\newblock In \emph{SC20: International Conference for High Performance Computing, Networking, Storage and Analysis}, pp.\  1--16. IEEE, 2020.

\bibitem[Shoeybi et~al.(2019)Shoeybi, Patwary, Puri, LeGresley, Casper, and Catanzaro]{megatron-lm}
Mohammad Shoeybi, Mostofa Patwary, Raul Puri, Patrick LeGresley, Jared Casper, and Bryan Catanzaro.
\newblock Megatron-lm: Training multi-billion parameter language models using model parallelism.
\newblock \emph{arXiv preprint arXiv:1909.08053}, 2019.

\bibitem[Simonyan et~al.(2013)Simonyan, Vedaldi, and Zisserman]{simonyan2013deep}
Karen Simonyan, Andrea Vedaldi, and Andrew Zisserman.
\newblock Deep inside convolutional networks: Visualising image classification models and saliency maps.
\newblock \emph{arXiv preprint arXiv:1312.6034}, 2013.

\bibitem[Tang et~al.(2024{\natexlab{a}})Tang, Sun, Li, Zhu, and Zhang]{tang2024logo}
Zecheng Tang, Zechen Sun, Juntao Li, Qiaoming Zhu, and Min Zhang.
\newblock Logo--long context alignment via efficient preference optimization.
\newblock \emph{arXiv preprint arXiv:2410.18533}, 2024{\natexlab{a}}.

\bibitem[Tang et~al.(2024{\natexlab{b}})Tang, Zhou, Li, Ji, Hou, and Zhang]{tang2024citeeval}
Zecheng Tang, Keyan Zhou, Juntao Li, Baibei Ji, Jianye Hou, and Min Zhang.
\newblock L-citeeval: Do long-context models truly leverage context for responding?
\newblock \emph{arXiv preprint arXiv:2410.02115}, 2024{\natexlab{b}}.

\bibitem[Tang et~al.(2025)Tang, Wang, Qiu, Ji, Sun, Zhou, Li, and Zhang]{tang2025loom}
Zecheng Tang, Haitian Wang, Quantong Qiu, Baibei Ji, Ruoxi Sun, Keyan Zhou, Juntao Li, and Min Zhang.
\newblock Loom-scope: a comprehensive and efficient long-context model evaluation framework.
\newblock \emph{arXiv preprint arXiv:2507.04723}, 2025.

\bibitem[Team et~al.(2024)Team, Georgiev, Lei, Burnell, Bai, Gulati, Tanzer, Vincent, Pan, Wang, et~al.]{team2024gemini}
Gemini Team, Petko Georgiev, Ving~Ian Lei, Ryan Burnell, Libin Bai, Anmol Gulati, Garrett Tanzer, Damien Vincent, Zhufeng Pan, Shibo Wang, et~al.
\newblock Gemini 1.5: Unlocking multimodal understanding across millions of tokens of context.
\newblock \emph{arXiv preprint arXiv:2403.05530}, 2024.

\bibitem[Wang et~al.(2023)Wang, Li, Dai, Chen, Zhou, Meng, Zhou, and Sun]{wang2023label}
Lean Wang, Lei Li, Damai Dai, Deli Chen, Hao Zhou, Fandong Meng, Jie Zhou, and Xu~Sun.
\newblock Label words are anchors: An information flow perspective for understanding in-context learning.
\newblock In \emph{Proceedings of the 2023 Conference on Empirical Methods in Natural Language Processing}, pp.\  9840--9855, 2023.

\bibitem[Wang et~al.(2025)Wang, Ding, Lv, Xu, Li, Shi, Zheng, and Huang]{wang2025layer}
Zhenghua Wang, Yiran Ding, Changze Lv, Zhibo Xu, Tianlong Li, Tianyuan Shi, Xiaoqing Zheng, and Xuanjing Huang.
\newblock Layer-specific scaling of positional encodings for superior long-context modeling.
\newblock \emph{arXiv preprint arXiv:2503.04355}, 2025.

\bibitem[Wolf et~al.(2020)Wolf, Debut, Sanh, Chaumond, Delangue, Moi, Cistac, Rault, Louf, Funtowicz, Davison, Shleifer, von Platen, Ma, Jernite, Plu, Xu, Scao, Gugger, Drame, Lhoest, and Rush]{wolf-etal-2020-transformers}
Thomas Wolf, Lysandre Debut, Victor Sanh, Julien Chaumond, Clement Delangue, Anthony Moi, Pierric Cistac, Tim Rault, Rémi Louf, Morgan Funtowicz, Joe Davison, Sam Shleifer, Patrick von Platen, Clara Ma, Yacine Jernite, Julien Plu, Canwen Xu, Teven~Le Scao, Sylvain Gugger, Mariama Drame, Quentin Lhoest, and Alexander~M. Rush.
\newblock Transformers: State-of-the-art natural language processing.
\newblock In \emph{Proceedings of the 2020 Conference on Empirical Methods in Natural Language Processing: System Demonstrations}, pp.\  38--45, Online, October 2020. Association for Computational Linguistics.
\newblock URL \url{https://www.aclweb.org/anthology/2020.emnlp-demos.6}.

\bibitem[Wu et~al.(2024)Wu, Wang, Xiao, Peng, and Fu]{wu2024retrieval}
Wenhao Wu, Yizhong Wang, Guangxuan Xiao, Hao Peng, and Yao Fu.
\newblock Retrieval head mechanistically explains long-context factuality.
\newblock \emph{arXiv preprint arXiv:2404.15574}, 2024.

\bibitem[Xi et~al.(2025)Xi, Huang, Liao, Huang, Guo, Liu, Zheng, Ye, Zhang, Chen, et~al.]{xi2025agentgym}
Zhiheng Xi, Jixuan Huang, Chenyang Liao, Baodai Huang, Honglin Guo, Jiaqi Liu, Rui Zheng, Junjie Ye, Jiazheng Zhang, Wenxiang Chen, et~al.
\newblock Agentgym-rl: Training llm agents for long-horizon decision making through multi-turn reinforcement learning.
\newblock \emph{arXiv preprint arXiv:2509.08755}, 2025.

\bibitem[Xiao et~al.(2024{\natexlab{a}})Xiao, Tang, Zuo, Guo, Yang, Tang, Fu, and Han]{xiao2024duoattention}
Guangxuan Xiao, Jiaming Tang, Jingwei Zuo, Junxian Guo, Shang Yang, Haotian Tang, Yao Fu, and Song Han.
\newblock Duoattention: Efficient long-context llm inference with retrieval and streaming heads.
\newblock \emph{arXiv preprint arXiv:2410.10819}, 2024{\natexlab{a}}.

\bibitem[Xiao et~al.(2024{\natexlab{b}})Xiao, Tian, Chen, Han, and Lewis]{xiaoefficient}
Guangxuan Xiao, Yuandong Tian, Beidi Chen, Song Han, and Mike Lewis.
\newblock Efficient streaming language models with attention sinks.
\newblock In \emph{The Twelfth International Conference on Learning Representations}, 2024{\natexlab{b}}.

\bibitem[Xu et~al.(2025)Xu, Xiao, Huang, Guo, and Han]{xu2025xattention}
Ruyi Xu, Guangxuan Xiao, Haofeng Huang, Junxian Guo, and Song Han.
\newblock Xattention: Block sparse attention with antidiagonal scoring.
\newblock \emph{arXiv preprint arXiv:2503.16428}, 2025.

\bibitem[Yang et~al.(2024)Yang, Yang, Zhang, Hui, Zheng, Yu, Li, Liu, Huang, Wei, et~al.]{yang2024qwen2}
An~Yang, Baosong Yang, Beichen Zhang, Binyuan Hui, Bo~Zheng, Bowen Yu, Chengyuan Li, Dayiheng Liu, Fei Huang, Haoran Wei, et~al.
\newblock Qwen2. 5 technical report.
\newblock \emph{arXiv preprint arXiv:2412.15115}, 2024.

\bibitem[Yang et~al.(2025)Yang, Li, Yang, Zhang, Hui, Zheng, Yu, Gao, Huang, Lv, et~al.]{yang2025qwen3}
An~Yang, Anfeng Li, Baosong Yang, Beichen Zhang, Binyuan Hui, Bo~Zheng, Bowen Yu, Chang Gao, Chengen Huang, Chenxu Lv, et~al.
\newblock Qwen3 technical report.
\newblock \emph{arXiv preprint arXiv:2505.09388}, 2025.

\bibitem[Ye et~al.(2024)Ye, Dong, Xia, Sun, Zhu, Huang, and Wei]{ye2024differential}
Tianzhu Ye, Li~Dong, Yuqing Xia, Yutao Sun, Yi~Zhu, Gao Huang, and Furu Wei.
\newblock Differential transformer.
\newblock \emph{arXiv preprint arXiv:2410.05258}, 2024.

\bibitem[Yen et~al.(2025)Yen, Gao, Hou, Ding, Fleischer, Izsak, Wasserblat, and Chen]{yen2025helmet}
Howard Yen, Tianyu Gao, Minmin Hou, Ke~Ding, Daniel Fleischer, Peter Izsak, Moshe Wasserblat, and Danqi Chen.
\newblock Helmet: How to evaluate long-context models effectively and thoroughly.
\newblock In \emph{The Thirteenth International Conference on Learning Representations}, 2025.

\bibitem[Yu et~al.(2024)Yu, Xu, and Akkiraju]{yu2024defense}
Tan Yu, Anbang Xu, and Rama Akkiraju.
\newblock In defense of rag in the era of long-context language models.
\newblock \emph{arXiv preprint arXiv:2409.01666}, 2024.

\bibitem[Yuan et~al.(2025)Yuan, Gao, Dai, Luo, Zhao, Zhang, Xie, Wei, Wang, Xiao, et~al.]{yuan2025native}
Jingyang Yuan, Huazuo Gao, Damai Dai, Junyu Luo, Liang Zhao, Zhengyan Zhang, Zhenda Xie, YX~Wei, Lean Wang, Zhiping Xiao, et~al.
\newblock Native sparse attention: Hardware-aligned and natively trainable sparse attention.
\newblock \emph{arXiv preprint arXiv:2502.11089}, 2025.

\bibitem[Zhang et~al.(2024{\natexlab{a}})Zhang, Bai, Lv, Gu, Liu, Zou, Cao, Hou, Dong, Feng, et~al.]{zhang2024longcite}
Jiajie Zhang, Yushi Bai, Xin Lv, Wanjun Gu, Danqing Liu, Minhao Zou, Shulin Cao, Lei Hou, Yuxiao Dong, Ling Feng, et~al.
\newblock Longcite: Enabling llms to generate fine-grained citations in long-context qa.
\newblock \emph{arXiv preprint arXiv:2409.02897}, 2024{\natexlab{a}}.

\bibitem[Zhang et~al.(2024{\natexlab{b}})Zhang, Hou, Lv, Cao, Hou, Niu, Hou, Dong, Feng, and Li]{zhang2024longreward}
Jiajie Zhang, Zhongni Hou, Xin Lv, Shulin Cao, Zhenyu Hou, Yilin Niu, Lei Hou, Yuxiao Dong, Ling Feng, and Juanzi Li.
\newblock Longreward: Improving long-context large language models with ai feedback.
\newblock \emph{arXiv preprint arXiv:2410.21252}, 2024{\natexlab{b}}.

\bibitem[Zhang et~al.(2024{\natexlab{c}})Zhang, Li, and Liu]{zhang2024extending}
Yikai Zhang, Junlong Li, and Pengfei Liu.
\newblock Extending llms' context window with 100 samples.
\newblock \emph{arXiv preprint arXiv:2401.07004}, 2024{\natexlab{c}}.

\bibitem[Zhao et~al.(2024{\natexlab{a}})Zhao, Wei, Zeng, Cheng, Yang, Cheng, Wang, Li, Wu, Zhu, et~al.]{zhao2024longskywork}
Liang Zhao, Tianwen Wei, Liang Zeng, Cheng Cheng, Liu Yang, Peng Cheng, Lijie Wang, Chenxia Li, Xuejie Wu, Bo~Zhu, et~al.
\newblock Longskywork: A training recipe for efficiently extending context length in large language models.
\newblock \emph{arXiv preprint arXiv:2406.00605}, 2024{\natexlab{a}}.

\bibitem[Zhao et~al.(2024{\natexlab{b}})Zhao, Yin, and Durrett]{zhao2024understanding}
Xinyu Zhao, Fangcong Yin, and Greg Durrett.
\newblock Understanding synthetic context extension via retrieval heads.
\newblock \emph{arXiv preprint arXiv:2410.22316}, 2024{\natexlab{b}}.

\end{thebibliography}
\bibliographystyle{iclr2026_conference}

\clearpage
\appendix
\section{Illustration of Training Efficiency of Current Methods}
\label{appdix:train_efficiency}
Training efficiency comparisons across current long-context methods are inherently challenging: performance gains typically exhibit diminishing returns with increased token budgets, and reported results often stem from divergent training setups — including data composition, optimization hyper-parameters, and hardware configurations. 
These factors render direct ``gain-per-token'' comparisons unreliable when conditions are unmatched.
To fairly compare training efficiency across methods — despite differing hyper-parameters and convergence behaviors — we adopt a controlled proxy: average task gain per 1B tokens, measured under identical data, optimizer, batch size, learning rate, and hardware (8 $\times$ A100 GPUs). 
Specifically, we compare ProLong~\citep{gao2024prolong} - one long-context SFT method, and LongCE~\citep{fang2024wrong} - one token-level re-weighting training method, on the Llama3-8B-Base model.
As shown in Table~\ref{tab:performance_comparison}, we evaluate model performance on LongBench-E (12 real-world tasks) per 50 training steps~(0.41B tokens per 50 steps), and find that LongCE achieves a 13-point gain per 1B tokens versus ProLong's 0.3-point gain per 1B tokens.

\begin{table}[ht]
\centering 
\small
\caption{Performance comparison between ProLong~(SFT) and LongCE across training steps, where each step contains the same training setting.}
\label{tab:performance_comparison}
\begin{tabular}{lccccc}
\toprule
Method & Step 0 & Step 50 & Step 100 & Step 150 & Step 200 \\
\midrule
ProLong (SFT) & 25.50 & 27.32 & 28.15 & 28.44 & 29.13 \\
LongCE (same data) & 25.50 & 28.30 & 29.72 & 31.01 & 32.91 \\
\bottomrule
\end{tabular}
\end{table}

\section{Preliminary Study Details}
\label{appdix:pre_task}

\subsection{Preliminary Task Construction}
\paragraph{Task Selection}
We select 3-hop and 4-hop tasks based on qa3 tasks in the BABILong Benchmark to build our datasets, as these tasks generally pose significant challenges for LLMs. 
However, it is worth noting that the original BABILong qa4 samples do not truly require 4-hop reasoning to produce correct outputs. 
For example, a sample from this subset with 0k context is shown in Figure~\ref{fig:qa4_sample}. 
In this case, the task only requires attention to a single fact, ``The bedroom is west of the bathroom'' to answer the question, while the first sentence serves as an interference fact. 
Even in terms of keywords, the model only needs to focus on three keywords: ``bathroom'', ``west'', and ``bedroom'' from the second sentence. 
Thus, we design our 4-hop dataset based on the BABILong qa3 source data, with one sample shown in Figure~\ref{fig:our_sample}. 
By carefully arranging the order of facts and reducing the conditions of questions in the long context, we ensure that the model is required to search for all four supporting facts in sequence to produce the correct output.

\begin{table}[h]
    \caption{Variable settings, where R. denotes random.}
    \label{tab:variables}
    \centering
    \begin{tabular}{c c c c}
    \toprule
     \bf Hops & \bf Samples  & \bf Permute & \bf Lengths \\
    \midrule 
    2  & 100 & 5 & 8K  \\
    \midrule 
    3/4 & R. & R. & 0k - 64k \\
    \bottomrule
    \end{tabular}
\end{table}

\paragraph{Controlled Evaluation Data Synthesis}
\label{appdix: build_data} 
We use the 4-hop task with non-zero context as an example here. 
As shown in Table~\ref{tab:variables}, all variables used for building data include the facts sample, the facts permutation, and the context length. 
Firstly, we select source samples from the BABILong official file ``qa3\_three-supporting-facts'' as our base data. 
Then, we modify the original BABILong qa3 supporting facts following the pattern shown in Figure~\ref{fig:qa4_pattern}. 
Afterward, we add interference to these four original facts while maintaining the relative order of the supporting facts. 
The process begins by selecting a noise context of the appropriate length and inserting the facts into it. 
Specifically, we divide the noise context into 10 equal-length chunks, leaving 10 candidate positions for the insertion of the 4 supporting facts (excluding the tail). 
Next, we randomly select five permutations from the full set of $C_{10}^{4}$ candidate position permutations. 
After injecting noise, we randomly insert interference facts, i.e., facts that are similar to the supporting facts but irrelevant, among all sentences. 
We ensure that at least one interference fact is placed after the last supporting fact to test the model's robustness. 
To ensure the correctness of the samples, we make sure that the objects appearing in the interference facts do not overlap with those in the supporting facts. 
Additionally, we ensure that the number of interference facts is between one and two times the number of supporting facts to avoid making the samples either too easy or too difficult. 
Finally, for all samples with the same context length, we use the same noise context to maintain consistency.
In the end, we randomly insert a few emojis into the constructed context to test the sensitivity of the model to low-frequency tokens.
For the 3-hop task, we directly use the original qa3 task format from BABILong as the base, and the subsequent processing follows a similar approach to the one described above for the 4-hop task.

\begin{table}[t]
    \caption{Performance statistics of using different numbers of attention heads on our preliminary synthetic task. Notably, we find that selecting the top-30 heads yields results that are nearly identical to those obtained when using all attention heads.}
    \centering
    \resizebox{\linewidth}{!}{
    \begin{tabular}{l c c c c c c c c}
    \toprule
    \multirow{2}{*}{\bf Head Number} & \multicolumn{2}{c}{\bf Supporting} & \multicolumn{2}{c}{\bf Interference} & \multicolumn{2}
    {c}{\bf Irrelevant} & \multicolumn{2}{c}{\bf Low-frequency} \\
    \cmidrule{2-9}
    & Correct & Wrong & Correct & Wrong & Correct & Wrong & Correct & Wrong \\
    \midrule
    Top-30 & 0.21 & 0.11 & 0.07 & 0.17 & 0.72 & 0.72 & 0.00	& 0.00 \\
    \midrule
    All	& 0.20 & 0.13 & 0.09 & 0.15 & 0.71 & 0.72 & 0.00 & 0.00 \\
    \bottomrule
    \end{tabular}}
    \label{tab:appdix_retrieval_head}
\end{table}

\begin{figure}[h]
\begin{AcademicBox}[\footnotesize One BABILong qa4 sample with 0k context]
Input :   
\begin{verbatim} 
The bedroom is west of the office. 
The bathroom is west of the bedroom.
\end{verbatim}
\vspace{-2pt} \hrule \vspace{4pt}
\textbf{\textit{Question:}} \\
What is west of the office? \\
\vspace{-5pt} \hrule \vspace{4pt}
\textbf{\textit{Supporting Facts: }} \\
 The bedroom is west of the office.\\
\vspace{-5pt} \hrule \vspace{4pt}
\textbf{\textit{Ground truth:}} \\
bedroom
\end{AcademicBox}
\caption{A BABILong qa4 sample with 0k context }
\label{fig:qa4_sample}
\end{figure}

\begin{figure}[t]
\begin{AcademicBox}[\footnotesize One of our 4-hop samples with 0k context]
Input :   
\begin{verbatim} 
Mary journeyed to the office. 
Mike went to the office.
Mary got the apple. 
Daniel picked up the football.
Daniel went back to the bedroom.
Mary journeyed to the bathroom. 
Mary dropped the apple.
Jonh went to the bathroom.
\end{verbatim}
\vspace{-2pt} \hrule \vspace{4pt}
\textbf{\textit{Question:}} \\
Where was the apple's location prior to the place where the apple was discarded, left or dropped? \\
\vspace{-5pt} \hrule \vspace{4pt}
\textbf{\textit{Supporting Facts:}} \\
 Mary journeyed to the office. \\
 Mary got the apple. \\
 Mary journeyed to the bathroom.\\
 Mary dropped the apple.\\
\vspace{-5pt} \hrule \vspace{4pt}
\textbf{\textit{Ground truth:}} \\
office
\end{AcademicBox}
\caption{One of our 4-hop samples with 0k context}
\label{fig:our_sample}
\end{figure}

\begin{figure}[t]
\begin{AcademicBox}[\footnotesize The pattern of our 4-hop sample]
\begin{verbatim} 
Supporting fact1: {x} {m} the {y1}   
Supporting fact2: {x} {p} the {o}
Supporting fact3: {x} {m} the {y2}
Supporting fact4: {x} {d} the {o} 
\end{verbatim}
\vspace{-2pt} \hrule \vspace{4pt}
\textbf{\textit{Question:}} \\
Where was the \textbf{\{o\}}'s location prior to the place where the \textbf{\{o\}} was discarded, left or dropped? 
\vspace{2pt} \hrule \vspace{4pt}
\textbf{\textit{Ground truth:}} \\
\textbf{\{y1\}}
\vspace{2pt} \hrule \vspace{4pt}
\textbf{\textit{Explanation:}} \\
\textbf{\{x\}} : a character name, selected from \{Mary, Daniel, Mike, ...\} \\
\textbf{\{m\}} : a predicate indicating movement, selected from \{went to, journeyed to, travelled to, ...\}\\
\textbf{\{y1\}, \{y2\}} : two different locations, selected from \{office, bedroom, bathroom, ...\}\\
\textbf{\{p\}} : a predicate indicating picking up, selected from \{picked up, took, grabbed, ...\}\\
\textbf{\{d\}} : a predicae indicating dropping, selected from \{dropped, put down, discarded, ...\}\\
\textbf{\{o\}} : an object name, selected from \{apple, football, milk, ...\}
\end{AcademicBox}
\caption{The pattern of our 4-hop sample}
\label{fig:qa4_pattern}
\end{figure}

% \begin{verbatim}
% {x} : a character name (selected from {Mary, Daniel, Mike,...})
% \end{verbatim}
% \begin{verbatim}
% {m} : a predicate indicating movement,
% (selected from {went to, journeyed to, travelled to ...})
% \end{verbatim}
% \begin{verbatim}
% {l} : a location (selected from {office, bedroom, bathroom, ...})
% \end{verbatim}
% \begin{verbatim}
% {p} : a predicate indicating picking up (selected from {picked up, took, grabbed, ...})
% \end{verbatim}
% \begin{verbatim}
% {d} : a predicae indicating dropping (selected from {dropped, put down, discarded, ...})
% \end{verbatim}
% \begin{verbatim}
% {o} : an object name (selected from {the apple, the football, the milk, ...})
% \end{verbatim}

\subsection{Design of IG Score}
\label{appdix:design_of_ig_score}
Prior work~\citep{wu2024retrieval} has shown that not all attention heads behave uniformly, i.e., some are specialized for retrieval-like behaviors, while others are not. However, it is important to note that these findings are primarily derived from studies focused on copy-oriented tasks, such as NIAH. In contrast, our task involves reasoning and inference, which fundamentally differs from the objectives of retrieval heads.
As a result, the mechanisms for attending to relevant context in our setting cannot be directly aligned with those used in retrieval-focused tasks.
To further analyze the appropriate number of attention heads to select, we conduct experiments where we select the top-k attention heads (k = 30) that retrieve the most relevant information based on the attention scores~(Table~\ref{tab:appdix_retrieval_head}). 
We find that the performance using only a subset of attention heads was highly consistent with the results obtained by averaging over all attention heads.
Therefore, for simplicity and ease of deployment, we adopt the latter approach, i.e., averaging IG scores across all attention heads.

\section{Derivation of Relation between Information Flow and Embedding Gradients}
\label{appedix:proof}
In transformer-based models, the Information Flow in attention is essentially the product of the attention distribution and its corresponding gradient.  
Therefore, we can transform the derivation into \textbf{constructing the gradient relationship between the attention score distribution~($A$) and the embedding~($E(X)$)}.  
This can be established via the chain rule and implemented through the specific computation steps of the attention mechanism.
Notably, in the following derivation, for simplicity, we omit the activation layers in the model. 
Additionally, considering that transformer-based models are composed of multiple identical network blocks stacked together, one can easily extend the conclusions from a single layer to multiple layers. 
Therefore, we focus on proving the case with \textbf{one embedding layer and one attention module}.

Given the basic definition of the attention mechanism, we have:
\begin{equation}
\left\{
\begin{array}{l l}
    Q = E(X) W_Q,  & A = \mathrm{softmax} \left(\frac{QK^{T}}{\sqrt{d}}\right), \\[5pt]
    K = E(X) W_K,  & O = A \cdot V, \\[5pt]
    V = E(X) W_V,  &
\end{array}
\right.
\nonumber
\end{equation}
where $W_{Q},W_{K},W_{V}\in \sR^{d\times d}$ are the model parameters, $O$ is the attention output, $E(X)\in \sR^{n\times d}$ is the input embedding matrix, $n$ and $d$ are sequence length and model dimension, respectively.

Let the loss function be $L$.
By the chain rule, the gradient of the loss with respect to $E(X)$ is:
\begin{align}
\nonumber
    \frac{\partial L}{\partial E(X)} = \frac{\partial L}{\partial O}\frac{\partial O}{\partial E(X)} = \frac{\partial L}{\partial A}\frac{\partial A}{\partial E(X)} \\+ \frac{\partial L}{\partial V}\frac{\partial V}{\partial E(X)}.
\end{align}
Since we have $\frac{\partial V}{\partial E(X)}=W_{V}^{T}$ and $\frac{\partial O}{\partial V}=A$, the gradient relationship between $A$ and $E(X)$ is:
\begin{align}
    \frac{\partial L}{\partial E(X)} \propto \frac{\partial L}{\partial A} \frac{\partial A}{\partial E(X)} 
    \label{equ:LE}
\end{align}
To eliminate the influence of the $\mathrm{Softmax(\cdot)}$ function, we can further decompose \eqref{equ:LE} into:
\begin{equation}
\left\{
\begin{aligned}
&S=\frac{QK^{T}}{\sqrt{d}},\\
&\frac{\partial L}{\partial E(X)} \approx \frac{\partial L}{\partial A}\cdot \left(\frac{\partial A}{\partial S}\cdot \frac{\partial S}{\partial E(X)}\right),    
\end{aligned}
\right.
\label{eq:2}
\end{equation}

where $\frac{\partial A}{\partial S}$ is the Jacobian of $\mathrm{Softmax}(\cdot)$ function, with elements $A_{ij}\left(\delta_{ik}-A_{ik}\right)$\footnote{$\delta_{ik}$ is the Kronecker delta function. If $i$ equals to $k$, $\delta_{ik}=1$, else $\delta_{ik}=0$. We can also rewrite this equation into $A_{ij}\left(1-A_{ij}\right)$.}.

For each element $S_{ij}=\frac{Q_{i}K_{j}^{T}}{\sqrt{d}}\in S$, the gradient with respect to $E(X)$ can be written as:
\begin{equation}
\begin{aligned}
\frac{\partial S_{ij}}{\partial E(X)} &= \frac{\partial\left(\frac{(E(X)_{i}W_Q)(E(X)_{j}W_K)^{T}}{\sqrt{d}}\right)}{\partial E(X)} \\&=\frac{1}{\sqrt{d}}\left(W_Q^{T}\cdot K_j  \cdot \delta_{ik}+ W_K^{T} \cdot Q_i \cdot \delta_{jk}\right).
\end{aligned}
\label{eq:3}
\end{equation}

Based on \eqref{eq:2} and \eqref{eq:3}, we can summary that:
\begin{equation}
\begin{aligned}
\frac{\partial L}{\partial E(X)_i} \propto &\underbrace{\frac{\partial L}{\partial A_{ij}}}_{\texttt{Sensitivity of $L$ to $A$}} \\& \times \underbrace{A_{ij}(1-A_{ij})}_{\texttt{Derivation from Softmax}} \\
&\times \underbrace{\frac{\partial S_{ij}}{\partial E(X)}}_{\texttt{Linear Transformation}}.
\end{aligned}
\label{eq:5}
\end{equation}

Based on \eqref{eq:5}, we can derive that when $A_{ij}$ increases, indicating higher attention between token $i$ and token $j$, the sensitivity of $L$ to $A$~($\frac{\partial L}{\partial A_{ij}}$) also increases. 
This results in larger derivatives on the embeddings. 
Additionally, if $A_{ij}$ becomes excessively large, approaching 1, the value of $A_{ij}(1-A_{ij})$ might tend toward 0. 
However, this is often not an issue in long-context scenarios, as the attention scores are unlikely to approach values near 0.5 due to the long context. 
Even if they exceed 0.5 (possibly for some special tokens), the increase in the first term~($\frac{\partial L}{\partial A_{ij}}$) helps mitigate this effect.

\section{Implementation Details}
\label{appdix:exp_details}

\subsection{Training Details}
For all experiments, we utilize the open-source training framework OpenRLHF\footnote{\url{https://github.com/OpenRLHF/OpenRLHF.git}}~\citep{hu2024openrlhf}, Ring-flash-attention\footnote{\url{https://github.com/zhuzilin/ring-flash-attention.git}}~\citep{liu2023ring} and DeepSpeed~\citep{rajbhandari2020zero}.
For LongCE training~\citep{fang2024wrong}, we set the sliding context window size as 8192 and employ the recommended hyper-parameters in the official code~\footnote{\url{https://github.com/PKU-ML/LongPPL.git}}.

\begin{table}[t]
    \centering
    \begin{minipage}{0.48\textwidth}
        \caption{Configuration of context window scaling training setting.}
        \resizebox{\linewidth}{!}{
        \begin{tabular}{l l}
        \toprule
        \multicolumn{2}{c}{Context Window Scaling Training Setting} \\
        \midrule
        Backbone & Llama-3-8B-base \\
        Training Objective & Language modeling \\
        RoPE base  & 20,000,000 \\
        Context window size & 8K $\rightarrow$ 64K \\
        Data seq-length & 64,000 \\
        Deepspeed & Zero2 \\
        Global batch size & 64 \\
        Epoch & 2 \\
        Training Steps & 160 \\
        Ring-attention size & 4 \\
        Learning-rate & 1e-5 \\
        LR-scheduler & cosine\_with\_min\_lr \\
        Optimizer & Adam~($\beta_1=0.9, \beta_2=0.95$) \\
        GPUs & A100~(80GB) $\times$ 8 \\
        Training time & $\approx$8h / epoch\\
        Training data & PG19~\citep{raecompressive2019} \\
        Total consumed tokens & 0.65B \\
        \bottomrule
        \end{tabular}}
        \label{tab:cws_config}
    \end{minipage}
    \hfill
    \begin{minipage}{0.48\textwidth}
        \caption{Configuration of language modeling training setting.}
        \resizebox{\linewidth}{!}{
        \begin{tabular}{l l}
        \toprule
        \multicolumn{2}{c}{Language Modeling Post-training Setting} \\
        \midrule
        Backbone & Llama-3.1-8B-base \\
        Training Objective & Language modeling \\
        RoPE base & 500,000 \\
        Context window size & 128K \\
        Data seq-length & 64,000 \\
        Deepspeed & Zero2 \\
        Epoch & 2 \\
        Global batch size & 32 \\
        Training Steps & 320 \\
        Ring-attention size & 4 \\
        Learning-rate & 1e-5 \\
        LR-scheduler & cosine\_with\_min\_lr \\
        Optimizer & Adam~($\beta_1=0.9, \beta_2=0.95$) \\
        GPUs & A100~(80GB) $\times$ 8 \\
        Training time & $\approx$8.5h / epoch \\
        Training data & PG19~\citep{raecompressive2019} \\
        Total consumed tokens & 0.65B \\
        \bottomrule
        \end{tabular}}
        \label{tab:lm_config}
    \end{minipage}
\end{table}

\begin{table}[t]
    \centering
    \begin{minipage}{0.48\textwidth}
        \caption{Configuration of long-context SFT training setting.}
        \resizebox{\linewidth}{!}{
            \begin{tabular}{l l}
            \toprule
            \multicolumn{2}{c}{Long-context Alignment Training Setting} \\
            \midrule
            Backbone & Llama-3.1-8B-Instruct \\
            Training Objective & Supervised fine-tuning \\
            RoPE base & 500,000 \\
            Context window size & 128K \\
            Data seq-length & 4,000$\sim$128,000 \\
            Deepspeed & Zero2 \\
            Global batch size & 32 \\
            Epoch & 2 \\
            Training Steps & 250 \\
            Ring-attention size & 4 \\
            Learning-rate & 1e-5 \\
            LR-scheduler & cosine\_with\_min\_lr \\
            Optimizer & Adam~($\beta_1=0.9, \beta_2=0.95$) \\
            GPUs & A100~(80GB) $\times$ 8 \\
            Training time & $\approx$6.5h / epoch\\
            Training data & \makecell[l]{LongMIT~\citep{chen2024essential}, \\LongAlpaca~\citep{long-alpaca}}  \\
            Total consumed tokens & 0.53B \\
            \bottomrule
        \end{tabular}}
        \label{tab:sft_config}
    \end{minipage}
    \hfill
    \begin{minipage}{0.48\textwidth}
        \centering
        \caption{Testing configuration of RULER}
        \resizebox{\linewidth}{!}{
            \begin{tabular}{l l}
            \toprule
            \multicolumn{2}{c}{Evaluation Configuration of RULER} \\
            \midrule
            Question Answering & qa\_1, qa\_2 \\
            Single NIAH & niah\_single\_1, \\ 
            & niah\_single\_2, \\ 
            & niah\_single\_3 \\
            Multi-keys NIAH & niah\_multikey\_1, \\ 
            & niah\_multikey\_2, \\ 
            & niah\_multikey\_3 \\
            Multi-values NIAH & niah\_multiquery \\
            Multi-queries NIAH & niah\_multivalue \\
            Others & common words extraction (CWE), \\ 
            & frequent words extraction (FWE), \\ 
            & variable tracking (VT) \\
            \midrule
            Length & 32K, 64K \\
            Num samples/task & 50 \\
            \bottomrule
        \end{tabular}}
        \label{tab:ruler_tasks_test_setting}
    \end{minipage}
\end{table}

\paragraph{Context Window Scaling}
To scale the context window size of the Llama-3-8B-base model from 8K to 64K~($8\times$), we adjust the RoPE base from 500,000 to 20,000,000 and directly train the model. We provide training configurations in Table~\ref{tab:cws_config}.

\paragraph{Data Post-processing Details}
For the context window scaling experiments, we employ the PG-19~\citep{raecompressive2019} dataset. 
For long-context SFT and CDT experiments, we construct our data from publicly available long-context QA datasets, including LongMiT~\citep{chen2024essential} and LongAlpaca~\citep{long-alpaca}.
The LongMiT dataset primarily consists of multi-hop QA tasks that require reasoning over 2 to 6 evidence passages. 
To adapt it for our setting, we apply two pre-processing steps: 
(i) Length distribution control — we constrain the sampled instances to fall within 16K–128K tokens. This range balances the need for sufficiently long contexts with training efficiency, given our compute resources (8 $\times$ A800 GPUs). Excessively long sequences were avoided as they considerably slow down training. 
(ii) Evidence balancing — we uniformly sample across different numbers of supporting passages to obtain a more balanced distribution for multi-hop reasoning.
To complement this, we include data from LongAlpaca, which predominantly features single-evidence QA with lengths around 16K tokens (under our model’s tokenizer). This addition enriches the training distribution by covering shorter single-evidence scenarios, which are underrepresented in LongMiT.
In total, our final training set comprises 7,000 samples from LongMiT and 1,000 samples from LongAlpaca, which are shuffled together before training.

\paragraph{Language Modeling Post-training and Long-context SFT}
The language modeling post-training and long-context SFT are directly applied to the Llama3.1-8B-base and Llama3.1-8B-Instruct, respectively, which already have 128K context window size.
We provide the training configurations in Table~\ref{tab:lm_config} and Table~\ref{tab:sft_config} respectively.

\begin{table}[t]
    \caption{Testing configuration of BABILong.}
    \centering
    \begin{tabular}{l c c c c c}
    \toprule
    \bf Metric & \bf QA1 & \bf QA2 & \bf QA3 & \bf QA7 & \bf QA8 \\
    \midrule
    Num & 100 & 100 & 100 & 100 & 100 \\
    Supporting Fact & 1 & 2 & 3 & 1$\sim$10 & 1$\sim$8 \\
    Interference Fact & 1$\sim$9 & 1$\sim$66 & 1$\sim$317 & 1$\sim$42 & 1$\sim$42 \\
    \bottomrule
    \end{tabular}
    \label{tab:BABILong_test_setting}
\end{table}

\begin{table}[t]
    \centering
    \caption{Evaluation results on HELMET~\citep{yen2025helmet}.}
    \small
    \begin{tabular}{l c c c c c c c c}
    \midrule
    Model & Recall & RAG & ICL & Re-rank & QA & Summ. & Cite & Avg. \\
    \midrule
    Claude-3.5-Sonnet & 94.7 &38.1& 61.0 &7.2 &12.6 &36.6 &18.7 & 38.4\\
    Mistral-Nemo-12B & 14.6 & 40.0&  84.0 & 0.0 & 22.5 & 18.5 & 0.5 & 25.7 \\ 
    ProLong-512K-Instruct & 98.8 & 63.2 & 86.5 & 22.5 & 43.9 & 29.2 & 1.4 & 49.4 \\
    Meta-Llama-3.1-8B & 95.2 &59.5 &83.9 &14.0 &43.2 &27.0 &2.9 &46.5 \\
    ~~+ CDT & 97.2 &  61.8 & 86.6 & 18.5 & 46.7 & 27.9 & 9.4 & \bf 49.7\\
    \bottomrule
    \end{tabular}
    \label{tab:helmet}
\end{table}

\begin{table}[t]
\centering
\caption{Evaluation results of two more LLMs on real-world long-context tasks and long-form reasoning tasks.}
\resizebox{\textwidth}{!}{
\begin{tabular}{l c c c c c c c c}
\toprule
\bf \multirow{2}{*}{Models} &\multicolumn{7}{c}{ \bf LongBench-E} & \bf BABILong \\
\cmidrule{2-9}
& \textbf{Type} & \textbf{S-Doc QA} & \textbf{M-Doc QA} & \textbf{Summ} & \textbf{Few-shot} & \textbf{Code} & \textbf{Avg.} & \bf Avg. \\
\midrule
Qwen2.5-7B-Instruct & - & 44.54 & 46.29 & 28.15 & 56.03 & 16.52 & 38.30	& 43.32 \\
~~~+ CDT & SFT & \bf 44.93 & \bf 47.29 & \bf 28.65 & \bf 57.33 & \bf 19.18 & \bf 39.48 & \bf 47.56 \\
\midrule
Qwen3-8B & - & 44.12 & 48.10 & 29.30 & 44.12 & 29.18 & 38.85 & 48.06 \\
~~~+ CDT & SFT & \bf 45.33 & \bf 49.13 & \bf 31.89 & \bf 46.24 & \bf 32.98 & \bf 41.11 & \bf 52.88 \\
\midrule
Mistral-V0.3-Instruct & - & 44.89 & 40.76 & 20.52 & 67.11 & 47.04 & 44.06 &  22.36   \\
~~~+ CDT & SFT & \bf 45.01 & \bf 41.79 &\bf  26.08 & \bf 67.75 & \bf 57.27 & \bf 47.58 & \bf 53.84 \\
\bottomrule
\end{tabular}}
\label{tab:more_res_LB}
\end{table}

\subsection{Evaluation Details}
\label{appdix:evaluation_results}
We conduct long-context evaluation mainly based on the long-context evaluation framework \texttt{LOOM-Scope}\footnote{\url{https://github.com/LCM-Lab/LOOM-Scope}}~\citep{tang2025loom}.

\paragraph{HELMET}
HELMET~\citep{yen2025helmet} is a comprehensive long-context evaluation benchmark containing 7 different subtasks, including recall, RAG, in-context learning~(ICL), re-rank, QA, summarization, and citation.
The context length of test samples ranges from 0 to 128K tokens. 
For inputs exceeding 128K, we truncate from the end to fit within the model's maximum context window.
We show the experimental results on HELMET in Table~\ref{tab:helmet}, where our CDT model achieves the best performance.

\paragraph{LongBench-E}
LongBench-E is a variant of LongBench~\citep{bai2024longbench} designed specifically for long-context real-world tasks.
We chose LongBench-E because it shares the same test dataset distribution as LongBench while covering a wider range of context lengths. 
For the Llama3-8B-base model, we truncate the input to 8K tokens, whereas for other models, we truncate the input to 32K tokens.

\paragraph{Language Modeling}
For the language modeling task, we calculate both LongPPL and PPL metrics on the GovReport dataset~\citep{huang-etal-2021-efficient}, which consists of long sequences from government reports. 
We sample 50 documents from GovReport, each with a context length of up to 32K tokens.

\paragraph{RULER}
RULER~\citep{hsieh2024ruler} is a comprehensive synthetic dataset that includes 6 different testing categories to evaluate a model's long-context understanding capabilities. 
We utilize all test categories, with each category containing 50 test samples covering lengths of 32K and 64K.
We post the testing configuration of RULER in Table~\ref{tab:ruler_tasks_test_setting}.

% \begin{table}[t]
%     \centering
%     \caption{Testing configuration of RULER}
%     \begin{tabular}{l l}
%     \toprule
%     \multicolumn{2}{c}{Evaluation Configuration of RULER} \\
%     \midrule
%     Question Answering & qa\_1, qa\_2 \\
%     Single NIAH & niah\_single\_1, \\ 
%     & niah\_single\_2, \\ 
%     & niah\_single\_3 \\
%     Multi-keys NIAH & niah\_multikey\_1, \\ 
%     & niah\_multikey\_2, \\ 
%     & niah\_multikey\_3 \\
%     Multi-values NIAH & niah\_multiquery \\
%     Multi-queries NIAH & niah\_multivalue \\
%     Others & common words extraction (CWE), \\ 
%     & frequent words extraction (FWE), \\ 
%     & variable tracking (VT) \\
%     \midrule
%     Length & 32K, 64K \\
%     Num samples/task & 50 \\
%     \bottomrule
%     \end{tabular}
%     \label{tab:ruler_tasks_test_setting}
% \end{table}

\paragraph{Long-form Reasoning}
We evaluate the long-form reasoning capability of models on selected tasks from BABILong~\citep{kuratov2024babilong}. 
Specifically, we select tasks that involve multiple supporting facts, as well as QA1, as the testing dataset. 
The BABILong testing configurations are shown in Table~\ref{tab:BABILong_test_setting}.

\subsection{Baseline Illustration}
We evaluate our method on three foundation models, i.e., LLaMA-3-8B-Base, LLaMA-3.1-8B-Base, and LLaMA-3.1-8B-Instruct—to ensure fair and consistent comparisons across all baselines. The baselines include: YaRN, which extends the context window using an improved NTK-based positional scaling method; CE (Cross Entropy), a standard language modeling objective without any context-aware weighting; LongCE, which builds upon the LongPPL method by identifying key tokens via perplexity during training and assigning them higher loss weights; SFT, an instruction tuning setup where input tokens are excluded from the loss calculation; and LOGO, a DPO-based training approach designed to mitigate misalignment in long-context tasks. 
\begin{wraptable}{r}{0.5\linewidth}
    \centering
    \caption{Model performance on language modeling tasks.}
    \label{tab:all_language_modeling}
    \begin{tabular}{lcc}
        \toprule
        \bf Models & \bf LongPPL & \bf PPL \\
        \midrule
        Llama-3-8B-Base & > 100 & > 100 \\
        ~~~+ YaRN & 3.55 & 5.60 \\
        ~~~+ CE & 3.90 & 6.46 \\
        ~~~+ LongCE & 3.55 & 5.60 \\
        ~~~+ CDT~(ours) & \bf 3.04 & \bf 5.40 \\
        \midrule
        Llama-3.1-8B-Base & 3.22 & \bf 4.79 \\
        ~~~+ CE & 3.28 & 4.86 \\
        ~~~+ LongCE & 3.24 & 5.28 \\
        ~~~+ CDT~(ours) & \bf 2.10 & 5.19 \\
        \midrule
        Llama-3.1-8B-Instruct & 4.05 & \bf 5.52 \\
        ~~~+ SFT & 3.31 & 5.51 \\
        ~~~+ LOGO & 4.11 & 5.54 \\
        ~~~+ CDT~(ours) & \bf 2.36 & 5.64 \\
        \bottomrule
    \end{tabular}
    \vspace{-4em}
\end{wraptable}
Additionally, we compare against several strong open-source long-context models: ProLong-512K-Instruct and NExtLong-512K-Instruct, which apply long-context scaling techniques on top of LLaMA-3-8B-Instruct and LLaMA-3.1-8B-Instruct, respectively; and LLaMA-3.1-8B-SEALONG, a DPO-trained model specifically optimized for long-context alignment.

\section{More Evaluation Results}
\label{appdix:more_eval_res}

\subsection{Analysis of Results on Real-world Long-context Tasks}
\label{appdix:cdt_scale_more_models}
The strong performance of CDT on code-related tasks, as shown in Table~\ref{tab:longbench}, is particularly notable. Code Completion requires models to accurately interpret local context and predict missing segments accordingly. CDT is especially well-suited for such tasks, as it enhances the model's ability to focus on local context information during generation, which likely contributes to the observed performance improvements.
Table~\ref{tab:case_study} offers a more intuitive illustration through a specific Code Completion example. In LongBench-E, this task is evaluated using the Edit Similarity~(Edit Sim) metric, which is highly sensitive to the number of tokens generated—especially under the official 64-token generation limit. In the provided example, LLaMA-3.1-8B-Instruct produces entirely incorrect outputs, while GPT-4o generates overly lengthy responses that negatively affect the Edit Sim score. In contrast, the CDT-enhanced model generates a concise and accurate response, resulting in a significantly higher Edit Sim score.
Furthermore, CDT leads to substantial improvements for both LLaMA-3.1-8B-Instruct and LLaMA-3-8B-Base on the Code task. These improvements can be attributed to two main factors. First, the training set includes code completion instances (e.g., 263 examples from LongMIT), which enable the model to learn relevant instruction-following patterns. Second, the baseline model's lower performance in this domain makes the gains from CDT more apparent. By contrast, LLaMA-3.1-8B-Base already demonstrates strong performance on code-related tasks—likely due to the composition of its pretraining data—resulting in smaller relative gains when CDT is applied.

\subsection{Generalizing CDT to More Models}
\label{appdix:cdt_scale_more_models}
We apply our CDT method to more LLMs, including Qwen2.5-7B-Instruct~\citep{yang2024qwen2} and Mistral-V0.3-Instruct~\citep{jiang2023mistral}.
We evaluation the model performance on real-world long-context tasks, long synthetic tasks, and long-form reasoning tasks.
We report the model performance in Table~\ref{tab:more_res_LB}, where we can observe that our CDT can significantly improve the model performance on different models.
For instance, the Mistral-V0.3-Instruct model obtains more than 30 points on the long-form reasoning task.

\subsection{Evaluation Results on Language Modeling Tasks}
Apart from evaluating with LongPPL on the language modeling task, we also calculate the PPL scores, which are shown in Table~\ref{tab:all_language_modeling}.

\subsection{Experiment Statistical Significance}
\label{appdix:exp_statistical_sign}
We collect the prediction results of the original model and the CDT model on the LongBench-E benchmark, and conduct a paired‑samples t-test to assess the statistical significance of the mean difference before and after the improvement, shown in Table~\ref{tab:significance}. The results show that our method significantly outperforms the baseline model at the 5\% significance level, indicating that our method achieves statistically significant improvements.

\begin{table}[t]
    \centering
    \caption{Statistical significance calculation on LongBench-E data with t‑Test.}
    \begin{tabular}{l l}
    \toprule
    Models & P-Value \\
    \midrule
    Llama3-8B-Base V.S. Llama3-8B-Base-CDT &  3.68e-15 \\
    Llama3.1-8B-Base V.S. Llama3.1-8B-Base-CDT &  1.53e-2\\
    Llama3.1-8B-Instruct V.S. Llama3.1-8B-Instruct-CDT & 2.39e-3  \\
    \bottomrule
    \end{tabular}
    \label{tab:significance}
\end{table}

\section{Analysis of Attention Map Before and After CDT}
\label{appdix:analysis_attn_map}
\begin{figure}[t]
    \centering
    \begin{subfigure}[b]{0.49\linewidth}
        \centering
        \includegraphics[width=\linewidth]{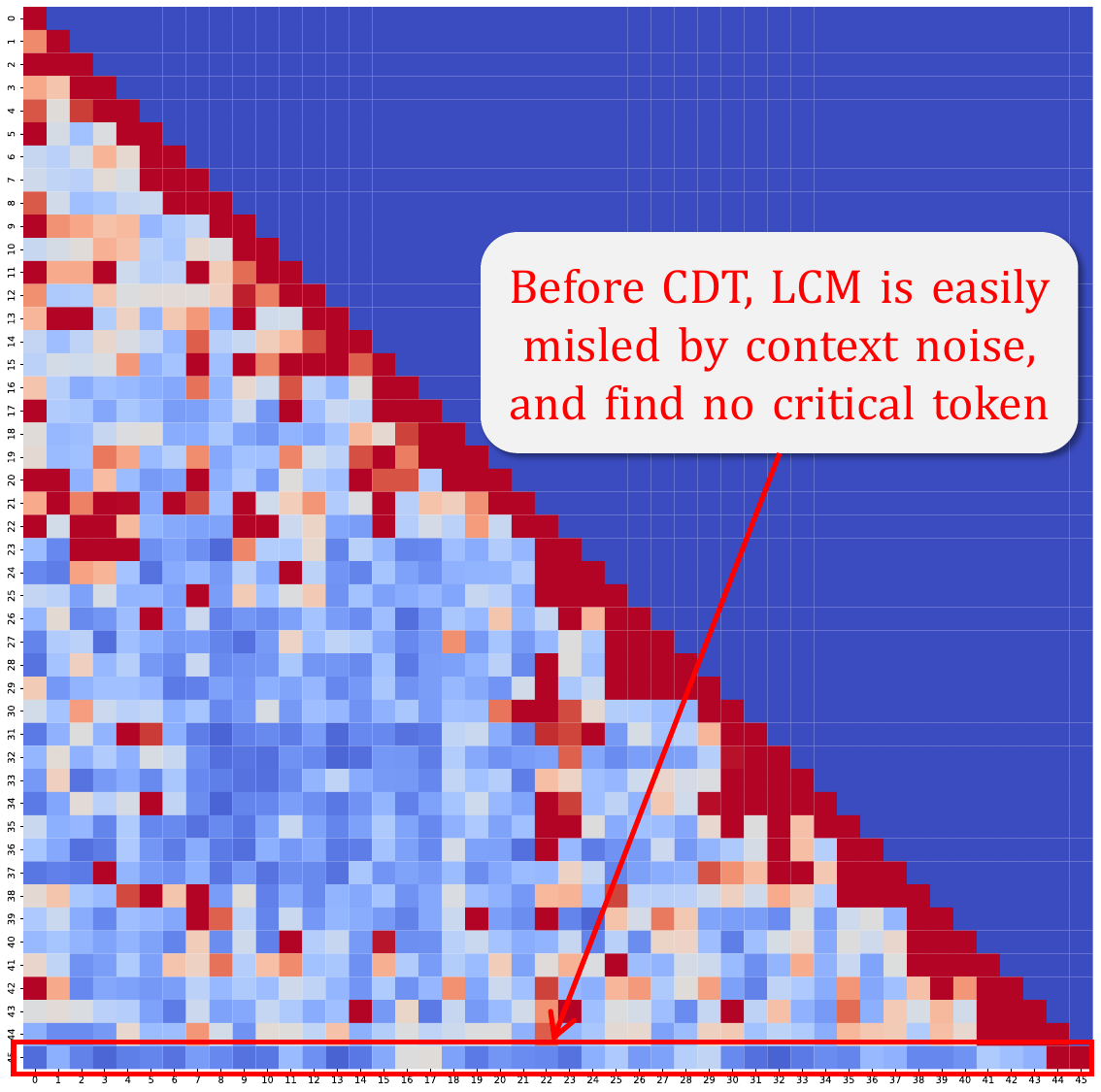}
        \caption{Attention map of 24th layer before CDT.}
        \label{fig:attn_map_before_cdt}
    \end{subfigure}
    \hfill
    \begin{subfigure}[b]{0.49\linewidth}
        \centering
        \includegraphics[width=\linewidth]{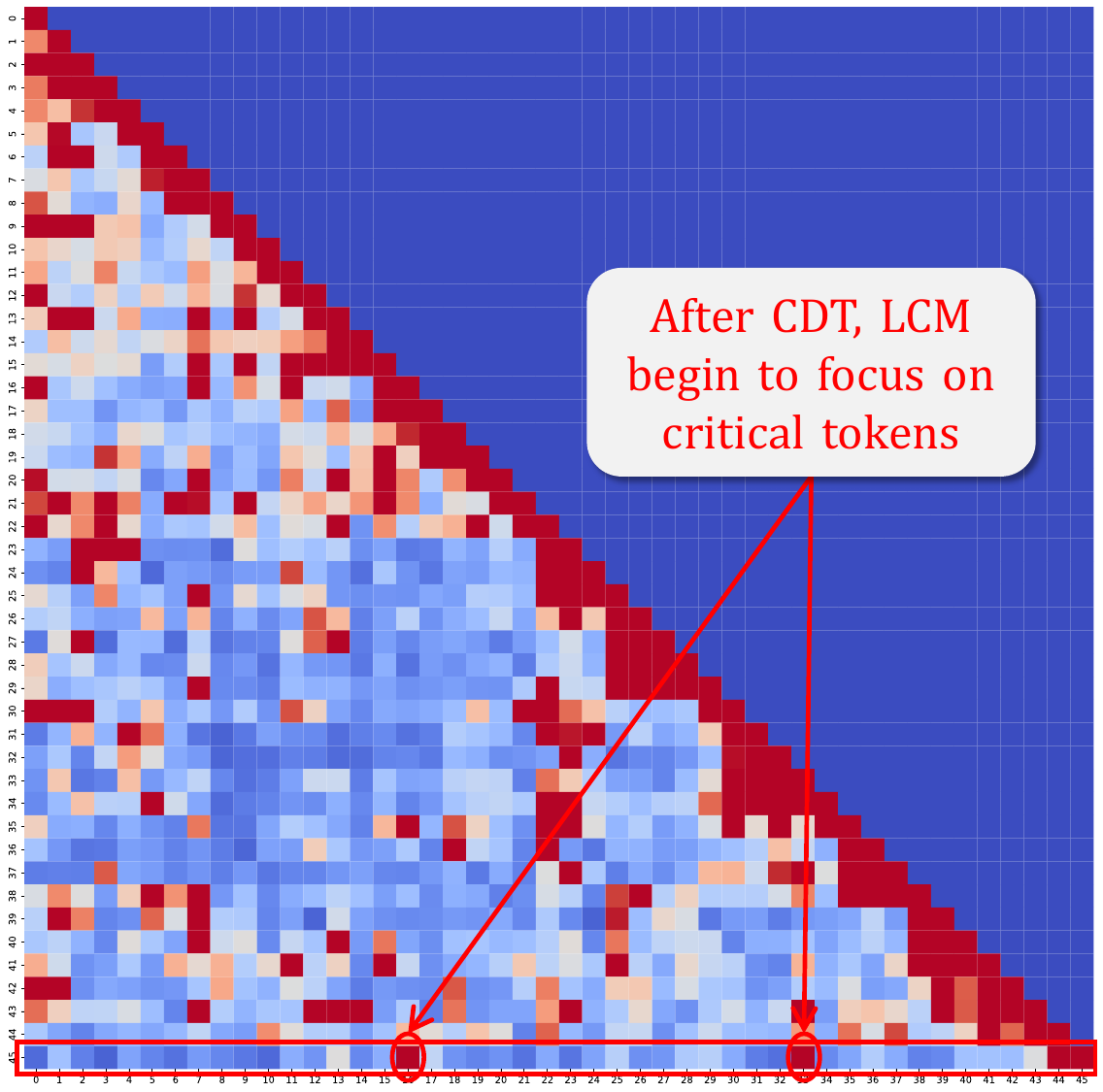}
        \caption{Attention map of 24th layer after CDT.}
        \label{fig:attn_map_after_cdt}
    \end{subfigure}
    \caption{Comparison between Attention Maps Before and After CDT. In each figure, a deeper (\textcolor{red}{red}) color indicates larger model attention to the corresponding context chunk. The final row of each map represents how the question attends to the entire input sequence, including both the context and the question itself. For clearer visualization, we recommend zoom in on this figure.}
    \label{fig:attn_map}
\end{figure}

In this section, we present a visualization of the model’s attention patterns before and after applying the CDT training strategy.
Given the long input length~(12,000 tokens) used in our evaluation, we evenly partition the input sequence into 46 chunks and calculate the total attention score for each chunk individually.
For each chunk, a higher total attention score indicates that the model places greater focus on this chunk. 
We visualize the attention maps of the 24th layer of the model, as this layer provides the clearest representation of CDT’s impact.
As shown in Figure~\ref{fig:attn_map}, we can observe that, before applying CDT, the model's attention is predominantly concentrated on the question itself~(the rightmost portion of the final row in Figure~\ref{fig:attn_map_before_cdt}), while key information within the context is largely overwhelmed by noise.
In contrast, after CDT training, the model not only attends to the question but also shows significantly increased attention to relevant contextual information, as highlighted by the red circles in the final row of Figure~\ref{fig:attn_map_after_cdt}.
\textbf{It is noteworthy that the attention map shows no significant changes before and after CDT training, indicating that CDT training does not compromise the original characteristics of the LCM.
Instead, it enhances the ability of LCM to capture critical information.}

\section{Limitation and Future Work}
\label{appdix:limitation}
Due to the expectation maximization~(EM) nature of CDT, it includes an additional context noise detection process, which introduces extra computational costs during the training phase. 
Although we have demonstrated in Section~\ref{subsec:step_performance} that these additional costs are negligible compared to the performance gains, theoretically, \textit{the noise detection cost will increase as the model size grows since it involves a complete forward-backward propagation process}. 
We leave this for future work, aiming to explore a simpler method for identifying the context noise or to develop more efficient model architectures. For example, designing specific network modules to handle noise, as proposed in \citet{ye2024differential}, could be a promising direction.
Additionally, we observe that \textit{the improvement brought by our method on complex reasoning tasks is not as significant as that on other tasks}, and we are yet to understand the relationship between this and the training data or the training objective function.
In the future, we aim to further investigate the impact of context noise on the model's long-form reasoning abilities, as well as the relationship between the CDT strategy and the enhancement of the model's reasoning capabilities.

\section{Use of LLMs}
\label{appdix:use_of_llm}
During the writing of this paper, we leveraged large language models (LLMs) to refine the clarity and fluency of our writing, particularly in the Abstract and Introduction sections. 
Specifically, we used the Qwen web interface~\footnote{\url{https://chat.qwen.ai}} to access the Qwen series of models (e.g., Qwen-Max), inputting early drafts of these sections and requesting stylistic improvements while preserving technical accuracy and original intent. 
The model’s suggestions helped enhance sentence structure, academic tone, and overall readability. All final content was carefully reviewed, validated, and edited by the authors to ensure fidelity to our research and adherence to scholarly standards.

\section{Error Analysis}
\label{appdix:error_analysis}
In this section, we analyze the error pattern of partial model predictions on real-world long-context tasks.
As shown in Table~\ref{tab:case_study}, we use colored text to highlight the correct and incorrect parts of the model's predictions.

\begin{table}[h]
    \caption{Generation samples of Llama3.1-8B-Instruct, Llama3.1-8B-Instruct-CD~(ours), and GPT4o on real-world long-context tasks.}
    \label{tab:case_study}
    \centering
    \begin{tabular}{p{0.48\linewidth} p{0.5\linewidth}}
    \toprule
    \multicolumn{2}{c}{Code Completion Task} \\
    \midrule
    \textbf{Question}: \texttt{Completing the code based on the context: [\colorbox{gray!40}{$\cdots$}]\textbackslash n self.interface = self.interface.replace(' ', '').lower() \textbackslash n self.param\_type = None \textbackslash n self.if\_type = None \textbackslash n \# state \textbackslash n self.results = dict() \textbackslash n [\colorbox{gray!40}{Completion}] self.result = dict() \textbackslash n def check\_params(self): \textbackslash n """Check all input params""" \textbackslash n if not self.interface: \textbackslash n self.module.fail\_json(msg='Error: Interface name cannot be empty.')[\colorbox{gray!40}{$\cdots$}]}
    & \textbf{Llama3.1-8B-Instruct}:  \texttt{\textcolor{red}{for (int i = 0; i < s.length(); i++) \{if (!Character.isDigit(s.charAt(i)) \&\& s.charAt(i) != ',') \}}} \newline
    \textbf{Llama3.1-8B-CD~(ours)}:  \texttt{\textcolor{deepgreen}{if self.interface == 'all':\textbackslash n} \textcolor{red}{self.param\_type = INTERFACE\_ALL\textbackslash n}} \newline
    \textbf{GPT4o}:  \texttt{\textcolor{deepgreen}{if self.interface.lower() == 'all':\textbackslash n} \textcolor{red}{self.param\_type = INTERFACE\_ALL\textbackslash n elif '/' in self.interface or self.interface.isdigit():\textbackslash n            self.param\_type = INTERFACE\_FULL\_NAME\textbackslash n        else:\textbackslash n            self.if\_type = get\_interface\_type(self.interface)\textbackslash n}} \newline
    \colorbox{green!50}{\textbf{Golden Answer}}:  \texttt{if self.interface == 'all':\textbackslash n}
    \\
    \midrule
    \multicolumn{2}{c}{Long-context Multi-hop Reasoning Task} \\
    \midrule
    \textbf{Question}: \texttt{What did M. King Hubbert's employer announce it was in the process of doing in April 2010? [\colorbox{gray!40}{$\cdots$}] In April 2010 Shell announced that the corporation is in process of trying to find a potential buyer for all of its operations in Finland and is doing similar market research concerning Swedish operations. [\colorbox{gray!40}{$\cdots$}]} & \textbf{Llama3.1-8B-Instruct}: \texttt{Shell announced the sale of some of its assets, including its liquefied petroleum gas (LPG) business, \textcolor{red}{to meet the cost of a planned \$28}} \newline
    \textbf{Llama3.1-8B-CD~(ours)}: \texttt{\textcolor{deepgreen}{trying to find a potential buyer for all of its operations in Finland} and is doing similar market research concerning Swedish operations.}\newline
    \textbf{GPT4o}: \texttt{MV Miralda's owner announced it was in the process of trying to \textcolor{deepgreen}{find a potential buyer for all of its operations} in Finland and in doing similar market research concerning.} \newline
    \colorbox{green!50}{\textbf{Golden Answer}}: \texttt{trying to find a potential buyer for all of its operations in Finland}
    \\
    \midrule
    \multicolumn{2}{c}{Single-QA Long-context Task} \\
    \midrule

    \textbf{Question}: \texttt{What metrics are used in evaluation?  [\colorbox{gray!40}{$\cdots$}] Figure FIGREF22 shows the results of these experiments wherein we are able to maintain the \textbf{precision} on the subset of the training set used for testing while reducing its \textbf{recall} and thus generalising the model. The fact that the model is generalizing on a dissimilar dataset is confirmed by the increase in the development set. \textbf{F1 score}.} [\colorbox{gray!40}{$\cdots$}] & \textbf{Llama3.1-8B-Instruct}:  \newline \texttt{\textcolor{red}{Macro-averaged F-score.}}  \newline
    \textbf{Llama3.1-8B-CD~(ours)}:\newline
    \texttt{\textcolor{red}{macro-averaged F-score, }\textcolor{deepgreen}{ precision, recall, F1 score.}} \newline
    \textbf{GPT4o}: \newline  \texttt{\textcolor{red}{Macro-averaged F-score.}} \newline
    \colorbox{green!50}{\textbf{Golden Answer}}: \newline \texttt{precision, recall , F1 score}
    \\
    \bottomrule
    \end{tabular}
\end{table}

\end{document}